\documentclass{article} \usepackage{heurevo_preprint,times}

\usepackage{amsmath,amsfonts,bm}

\def\eqref#1{equation~\ref{#1}}

\def\1{\bm{1}}

\DeclareMathAlphabet{\mathsfit}{\encodingdefault}{\sfdefault}{m}{sl}
\SetMathAlphabet{\mathsfit}{bold}{\encodingdefault}{\sfdefault}{bx}{n}

\usepackage{hyperref}
\usepackage{algorithm}
\usepackage{algpseudocode}

\usepackage{tabularx}

\usepackage{amsmath}
\usepackage{amssymb}
\usepackage{mathtools}
\usepackage{amsthm}
\usepackage{setspace}

\usepackage[table,dvipsnames]{xcolor}
\usepackage{graphicx}
\usepackage{booktabs}
\usepackage[T1]{fontenc}
\usepackage{makecell}
\usepackage{xspace}
\usepackage{enumitem}
\usepackage{tcolorbox}
\usepackage{listings}
\usepackage{multirow}

\usepackage{subfig}

\usepackage{wrapfig}

\usepackage{caption}
\usepackage[utf8]{inputenc}
\usepackage{longtable}
\usepackage{array}
\usepackage{float}
\usepackage{xurl}
\usepackage{setupboxes}

\usepackage{pifont}

\tcbuselibrary{skins,breakable,listings}

\newcommand{\algo}{\ensuremath{\text{HeurEvo}}\xspace}

\renewcommand{\cite}{\citep}

\title{HeurEvo: Agentic Evolution of Hybrid\\ Solver-Augmented Heuristics for\\ Time-Critical Mathematical Optimization}

\author{Feijie Wu\textsuperscript{1,}\thanks{Work done during an internship at Microsoft.}\quad Hugo Barbalho\textsuperscript{2}\quad Konstantina Mellou\textsuperscript{2}\\[4pt]
Marco Molinaro\textsuperscript{2}\quad Jing Gao\textsuperscript{1}\quad Ishai Menache\textsuperscript{2}\\[4pt]
Xinzhi Zhang\textsuperscript{2}\quad Sirui Li\textsuperscript{2}\\[7pt]
{\small\textsuperscript{1}Purdue University\quad
\textsuperscript{2}Microsoft Research}}
\date{}
\hypersetup{
  pdftitle={HeurEvo: Agentic Evolution of Hybrid Solver-Augmented Heuristics for Time-Critical Mathematical Optimization},
  pdfauthor={Feijie Wu, Hugo Barbalho, Konstantina Mellou, Marco Molinaro, Jing Gao, Ishai Menache, Xinzhi Zhang, Sirui Li},
  pdfsubject={Agentic heuristic design for mathematical optimization},
  pdfkeywords={heuristic design, program evolution, mathematical optimization},
  hidelinks
}

\begin{document}

\maketitle

\begin{abstract}

Recent advances in agentic heuristic design use AI agents and execution feedback to automate algorithm discovery for challenging optimization problems. In many practical settings, high-quality solutions must be obtained under strict runtime constraints, motivating hybrid approaches that combine problem-specific heuristics with powerful mathematical programming solvers. However, existing approaches typically improve heuristic components within predefined procedures or tune solver configurations in isolation. This limits holistic adaptation of where to allocate computation, how to leverage solvers, and how to refine the overall algorithmic structure. To address these limitations, we propose HeurEvo, an \textit{automated plan--code--component co-evolution framework} that jointly evolves the high-level algorithmic structures, their implementations, and a shared pool of reusable components. A planner determines which algorithmic components to use, how to combine them, and how to allocate runtime across stages, a coder realizes the resulting plan as executable code, while a component evolver updates the shared component pool. Within an island-based evolutionary framework, plans and implementations co-evolve with feedback from an interpreter agent that analyzes execution results and identifies opportunities for improvement. Across diverse combinatorial optimization benchmarks and challenging MIPLIB instances, HeurEvo finds high-quality solutions within tight runtime budgets, often matching or surpassing state-of-the-art optimization solvers given hours or days of computation. On several nonlinear geometry problems such as hexagon packing, it also improves upon the best previously reported results. These results highlight the value of jointly searching over algorithmic structure and implementation for agentic heuristic design.

\end{abstract}

\section{Introduction}

Mathematical optimization supports decisions in production scheduling, delivery routing, and resource allocation, where a solution is useful only if it can be obtained in time. Under a short runtime budget, a practical approach is to combine problem-specific heuristics and mathematical programming solvers in a multi-step algorithm \cite{talbi2002taxonomy,fischetti2003local}. Heuristics can quickly construct or repair solutions, while a solver can optimize selected subproblems or polish a promising incumbent within the broader search \cite{danna2005exploring}. Such hybrid algorithms have proved useful in practical applications, such as industrial production planning \cite{lee2023matheuristic}. Their success, however, depends on designing a complete solver-augmented algorithm suited to both the problem and its available solving time.

Recent methods use AI agents to automate heuristic and algorithm discovery, as well as solver configuration tuning. However, prior automatic heuristic-design methods typically restrict their search to individual components, such as scoring or update rules, within a fixed algorithmic framework such as ant colony optimization (ACO) or guided local search (GLS) \cite{liu2024evolution,ye2024reevo,wu2026refineevo}. Solver-oriented work, in turn, searches over solver configurations while retaining the solver's overall procedure \cite{luo2026grimip}. 
General-purpose code-evolution methods \cite{cemri2026adaevolve,liu2026evox,openevolve} broaden the search space by modifying entire programs, but they do not explicitly separate high-level algorithm structure from low-level implementation.
This leaves the design of complete hybrid optimization algorithms underexplored, especially how to combine heuristic and solver methods and allocate a shared runtime budget among them, which motivates our research question:

\emph{How can agents jointly evolve the composition and implementation of hybrid optimization algorithms to obtain better solutions within a fixed execution-time budget?}

We answer this question with \algo{}, an automated \textit{plan--code--component co-evolution framework} that jointly evolves high-level algorithmic plans, their executable implementations, and a shared pool of reusable components. A planner constructs and revises plans by selecting components, ordering their subgoals, and allocating runtime across stages. A coder implements each plan and refines its code and runtime allocation, while the planner revisits the algorithmic composition when code-level improvements stall. A component evolver uses experience accumulated during search to refine the shared component pool, allowing subsequent plans to build on improved components. An interpreter analyzes execution results and step traces to guide this evolution. Within an island-based evolutionary framework, different plans and their implementations evolve in separate populations while sharing knowledge through the component pool.

\begin{figure}[t]
\centering
\resizebox{\textwidth}{!}{\includegraphics[height=4cm]{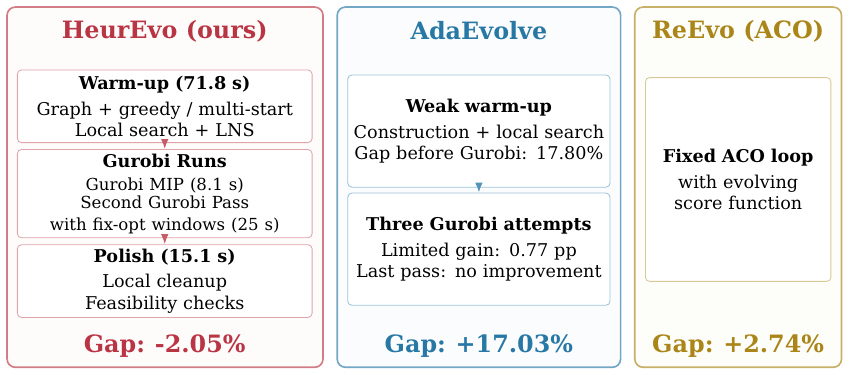}\hspace{0.25cm}\includegraphics[height=4cm]{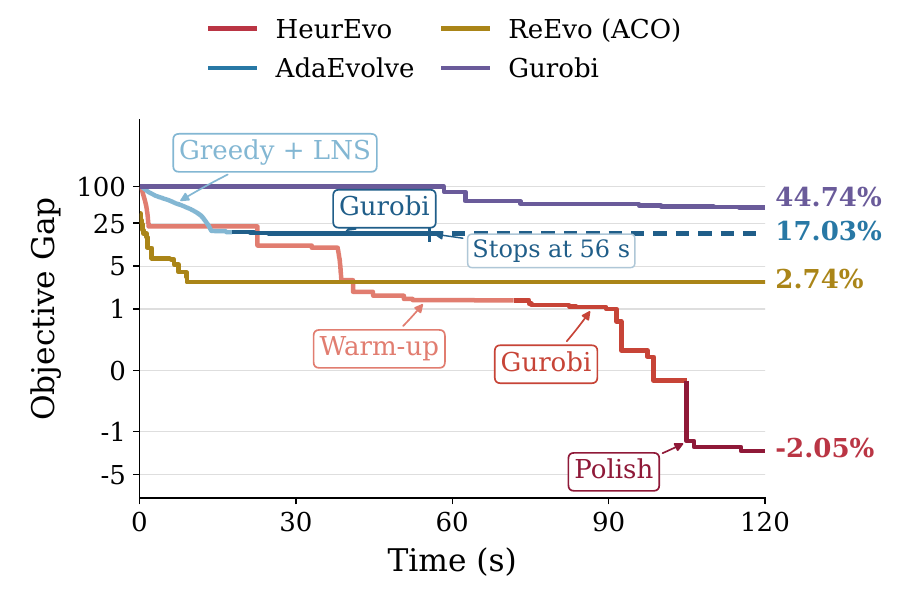}}
\caption{Comparison of heuristics generated from HeurEvo, AdaEvolve, and ReEvo (ACO) for an orienteering problem. \textbf{Left:} Stages (or steps) of the selected heuristic programs; bold values at the bottom report their measured objective gaps (lower is better). \textbf{Right:} Objective-gap trajectories on the same orienteering instance. AdaEvolve's program remains inactive after 52.2 seconds.}
\label{fig:op_case_analysis}
\end{figure}

Figure~\ref{fig:op_case_analysis} illustrates the resulting behavior. The evolved hybrid algorithm first combines greedy construction and large neighborhood search (LNS) to obtain a promising solution, then uses Gurobi to optimize selected subproblems, followed by solver-based polishing and local cleanup. By jointly choosing these components, their ordering, and their runtime allocation, \algo{} makes effective use of the available execution budget and achieves a lower objective gap than the illustrated programs produced by whole-program evolution and ReEvo's fixed ACO framework with a single LLM-evolved subcomponent, which is the heuristic function guiding node selection \cite{ye2024reevo}.

Our goal is to discover fast hybrid heuristics that generalize across related problem instances, producing high-quality solutions within a tight execution budget (two minutes in our experiments) without rerunning the discovery process for each new instance. We evaluate \algo{} on 11 synthetic tasks, six MIPLIB-NL-derived problems \cite{li2026constructing}, and four nonlinear geometry families. On the synthetic tasks, \algo{} achieves mean gaps of $-2.22\%$ on training instances and $-1.45\%$ on held-out test instances relative to Gurobi's 48-hour incumbents \cite{gurobi}, outperforming these incumbents on eight training tasks and seven test tasks. On the MIPLIB-NL problems, \algo{} improves upon AdaEvolve on four of six training problems. For nonlinear geometry, combining a \algo{} phase with subsequent AdaEvolve refinement produces several objectives numerically better than the reported reference values. Ablation studies further support the benefit of combining adaptive plan revision with component evolution.

\paragraph{Our contributions.}
\begin{itemize}
[topsep=0pt,itemsep=2pt,parsep=0pt,partopsep=0pt,leftmargin=*]

\item \textbf{Structured search over complete hybrid algorithms.}
We formulate automated heuristic design as an explicit search over complete solver-augmented algorithms. An evolvable plan selects heuristic and solver components, specifies their execution order and subgoals, and allocates a shared runtime budget across them. This makes algorithm composition a first-class search variable.

\item \textbf{Plan--code--component co-evolution.}
We introduce \algo{}, a plan--code--component co-evolution framework that jointly evolves algorithmic plans, their executable implementations, and a shared pool of reusable components. Code evolution improves implementations under the current plan; adaptive plan review revises the algorithmic structure when code-level progress stalls; and component evolution refines reusable components using execution experience accumulated across islands. Together, these mechanisms allow experience from previous executions to improve both current algorithms and future planning.

\item \textbf{Strong empirical performance under tight runtime budgets.}
Across synthetic combinatorial optimization, MIPLIB-NL-derived, and nonlinear geometry benchmarks, \algo{} achieves strong performance with a two-minute execution budget, often matching or improving upon far longer-running solver baselines and prior code-evolution methods. It also contributes to new best reported results on several nonlinear geometry problems, with ablations supporting the value of adaptive plan review and component evolution.

\end{itemize}

\section{Related Work}

\paragraph{Island-based evolutionary program search.}
LLM-guided evolutionary program search has emerged as a general paradigm for automated algorithm discovery, from seminal systems such as FunSearch \cite{romera2024mathematical} and AlphaEvolve \cite{novikov2025alphaevolve} to a growing set of follow-up frameworks \cite{lehman2023evolution,lange2026shinkaevolve,jiang2026deltaevolve,wang2025thetaevolve,assumpccao2025codeevolve,khrulkov2025gigaevo,wan2025loongflow}. At their core, these frameworks use LLMs to generate candidate programs and execution feedback to iteratively improve them through population-based search. Island-based variants extend this process across multiple semi-independent populations, preserving diverse search trajectories and reducing premature convergence. Notably, OpenEvolve \cite{openevolve} provides a competitive open-source implementation that combines MAP-Elites \cite{mouret2015illuminating}, exploratory and elite parent sampling, and migration across islands. AdaEvolve \cite{cemri2026adaevolve} and EvoX \cite{liu2026evox} further extend this paradigm: AdaEvolve adaptively controls exploration and evaluation allocation while introducing high-level tactics to redirect stalled search, whereas EvoX co-evolves candidate solutions with search policies that determine which candidates to build on and how new variants are generated. 

Our work differs primarily in the object being evolved. Rather than treating the candidate primarily as a program, \algo{} explicitly represents a hybrid optimization algorithm as a composition of heuristic and solver components together with an executable implementation. It then co-evolves the components, their composition into a plan, and the corresponding code, allowing execution feedback to revise both algorithmic structure and implementation.

\paragraph{Automatic Heuristic Design for Optimization Problems.}
A growing body of work applies LLM-based automatic heuristic design to optimization problems \cite{liu2023algorithm,van2024llamea,dat2025hsevo,yao2025multi,yang2025heuragenix,kuang2026llm}. Early approaches such as FunSearch \cite{romera2024mathematical} and EoH \cite{liu2024evolution} demonstrated that LLMs can iteratively discover effective optimization heuristics through program generation and execution feedback. ReEvo \cite{ye2024reevo} augments this process with short- and long-term reflection, MCTS-AHD \cite{zheng2025monte} uses Monte Carlo tree search to explore promising heuristic lineages, and RefineEvo \cite{wu2026refineevo} adaptively selects and refines evolutionary operators based on search progress. More recent work broadens the unit of evolution beyond a single heuristic component. MILP-Evo \cite{nie2026milp} jointly evolves cut-selection and branching rules within a branch-and-cut solver. MOTIF \cite{kiet2026motif} jointly optimizes multiple interacting algorithmic components, while MuEvo \cite{lv2026muevo} co-evolves heuristic ensembles and adaptively manages their constituent components. These methods substantially expand the search space, but operate within predefined algorithmic roles or solver structures. At the other extreme, concurrent work such as ATLAS \cite{yazdani2026atlas} evolves entire executable programs, allowing broad structural changes but leaving algorithm composition implicit in the code.

\algo{} instead makes the structure of a complete hybrid optimization algorithm an explicit object of evolution. Its plan can select heterogeneous heuristic and solver components, organize them into ordered steps with explicit subgoals, and allocate a shared runtime budget across them, while code evolution separately refines how each step is implemented. Component evolution further updates the reusable building blocks from which future plans are constructed.

\section{Method} \label{sec:method}

\subsection{Problem Formulation and Hybrid Algorithm Composition}
\label{sec:hierarchy}

\paragraph{Problem formulation.} Let $\mathcal{P}$ denote a distribution over instances of a target optimization problem family. We construct a training set $\mathcal{D}=\{d_j\}_{j=1}^{N}$ of $N$ instances, where $d_j\sim\mathcal{P}$.

Given $\mathcal{D}$ and a per-instance runtime budget $B$, our objective is to discover an executable program that achieves high average solution quality. We represent a hybrid algorithm $G$ as an ordered composition of $m$ heuristic or solver operations, i.e.  $G = g_1 \odot g_2 \odot \cdots \odot g_m$, where $\odot$ denotes sequential composition and intermediate results are passed between steps. Let $T_i(G;d)$ denote the runtime of each operation $g_i$ when executed as part of $G$ on instance $d$, including its associated overhead, and let $B$ denote the total runtime budget. Furthermore, let $f(G;d)$ denote the fitness score of the returned solution, with larger values indicating better solution quality. Under a fixed execution environment, we formulate the program search problem as
\begin{align}
    \max_{G} \quad
    F(G)=\frac{1}{|\mathcal{D}|}
      \sum_{d\in\mathcal{D}} f(G;d)
    \qquad \text{s.t.}\qquad
    T(G;d)=\sum_{i=1}^{m}T_i(G;d)\leq B,
      \;\; \forall d\in\mathcal{D}.
    \label{eq:constraint}
\end{align}

\paragraph{Component, plan, and code.}
A \emph{component} is a reusable high-level optimization method that admits multiple concrete implementations, such as greedy construction, local search, or calls to a mathematical programming solver. The \emph{component library} contains a set of such reusable components. A \emph{plan} specifies how these components are composed into an algorithm. It selects and orders a sequence of components and, for each component, specifies (1) a \emph{subgoal}, i.e., the intermediate outcome the component should achieve; (2) implementation guidance, such as the neighborhood used for local search; and (3) a time budget. We refer to a component together with these specifications as a \emph{step}. Since steps are executed sequentially, we index them from 1 to the total number of steps.

Executable \emph{code} instantiates the plan as an algorithm $G$, passing intermediate results between steps while respecting the total per-instance runtime budget $B$. Code evolution may modify implementation guidance and per-step time allocations while preserving the \emph{plan structure}, defined as the ordered sequence of components and their associated subgoals. Changes to this structure require plan evolution. Table~\ref{tab:example-plan} shows an example plan produced during an actual traveling salesperson problem (TSP) run.

\begin{table}[H]
\centering
\setlength{\belowcaptionskip}{3pt}
\caption{An example plan for the traveling salesperson problem.}
\label{tab:example-plan}
\footnotesize
\setlength{\tabcolsep}{2.5pt}
\renewcommand{\arraystretch}{1.02}
\setlength{\extrarowheight}{0.1em}
\resizebox{0.8\linewidth}{!}{\begin{tabularx}{\linewidth}{@{}>{}c >{\raggedright\arraybackslash\bfseries}p{0.21\linewidth} >{\raggedright\arraybackslash}p{0.20\linewidth} >{\raggedright\arraybackslash\itshape}X >{\raggedright\arraybackslash}p{0.13\linewidth}@{}}
\toprule
Step & Component & Subgoal & Implementation details & Time budget \\
\midrule
1 & Euclidean tour construction & Produce feasible and diverse incumbents & Build a small portfolio of nearest-neighbor and insertion tours; retain a diverse elite set & $9$ s \\
2 & Local tour improvement & Rapidly reduce tour length with cheap moves & Apply candidate-restricted 2-opt and Or-opt descent; perturb and restart only after stagnation & $20$ s \\
3 & Candidate-edge restricted MIP & Escape local minima through nonlocal edge changes & Warm-start Gurobi on a sparse graph containing proximity, incumbent, elite, and repair arcs & $32$ s \\
4 & Large-neighborhood search & Repair costly regions missed by the sparse global search & Fix most of the incumbent, destroy a long-edge or geographic region, and use Gurobi to reconnect that region exactly & $42$ s \\
5 & Residual anytime primal portfolio & Use the residual budget on the most productive search mode & Adaptively select among Gurobi LNS micro-repair, Gurobi sparse-MIP retry, and heuristic local-search restart using same-run progress & Remaining time  \\
\bottomrule
\end{tabularx}
}
\end{table}

\subsection{Our Plan-Code-Component Co-Evolution Framework}
\label{sec:coevolution}

The representation above makes explicit three coupled design decisions: which components to use for a given problem, how to compose them, and how to implement each step under a shared runtime budget. To address these coupled decisions, we introduce \algo{}, a \emph{plan--code--component co-evolution framework} that jointly evolves the components themselves, the plans that compose them, and the code that implements those plans.

\begin{figure}[t]
    \centering
    \includegraphics[width=0.9\linewidth]{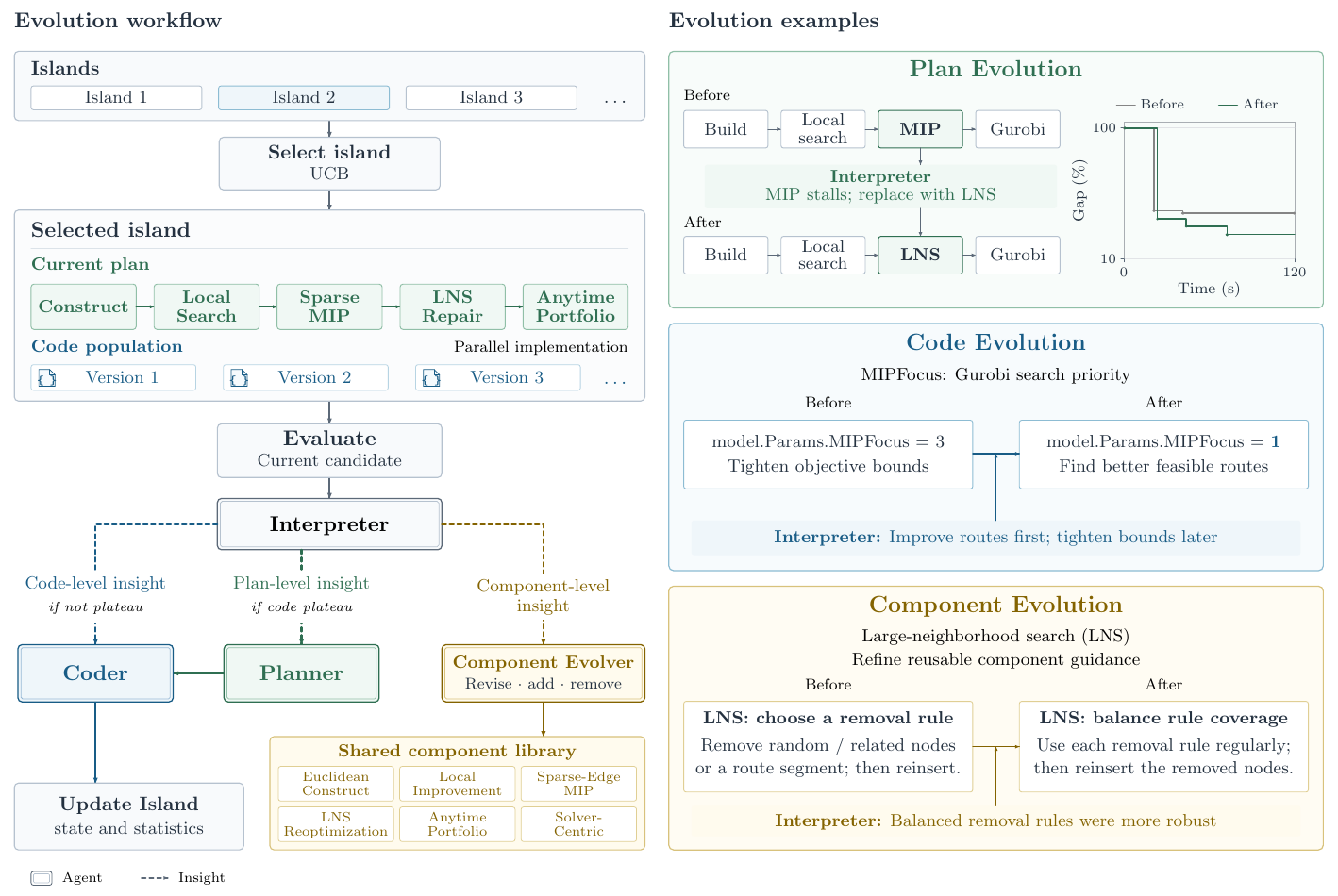}
    \caption{\algo workflow and examples of its three levels of evolution. \textbf{Left:} Simplified workflow. UCB selects an island for code evaluation; Interpreter feedback guides code refinement, plan revision when progress stalls, and updates to the shared component library. \textbf{Right:} Example edits. \emph{Top:} Plan evolution replaces a stalled mixed-integer programming (MIP) step with large-neighborhood search (LNS). \emph{Middle:} Code evolution changes Gurobi's \texttt{MIPFocus} setting while preserving the plan structure. \emph{Bottom:} Component evolution refines reusable LNS guidance to encourage balanced use of removal rules.}
    \label{fig:framework}
\end{figure}

To structure this search, we initialize a shared \emph{component library} and generate alternative plans from it. Each plan defines a plan-conditioned \emph{island} whose programs share the same ordered subgoals and selected components. Islands maintain separate active code populations and plan-specific feedback while sharing the component library and an archive of evaluated programs. Figure~\ref{fig:framework} (left) summarizes the resulting evolution workflow.

The examples in Figure~\ref{fig:framework} (right) illustrate three complementary levels of evolution:
\begin{itemize}[topsep=0pt,itemsep=0pt,parsep=0pt,partopsep=0pt,leftmargin=*]
\item \textbf{Component-level evolution} aims to refine the high-level search space. Based on accumulated feedback and evaluation evidence, the \emph{component evolver} revises existing component descriptions, introduces previously absent components, and removes components that remain ineffective across evaluated plans and implementations.
\item \textbf{Plan-level evolution} aims to discover more effective compositions of heuristics and solver operations when code-level improvements plateau. It may add, remove, replace, or reorder steps, or change their subgoals or selected components. The \emph{planner} considers programs under the current plan, the performance of plans on other islands, and feedback accumulated from code-level evolution on the current island. It proposes a revised plan with preliminary implementation procedures, which the Plan-based coding agent (coder) translates into a new executable code.
\item \textbf{Code-level evolution} aims to improve solution quality through better step implementations under a fixed plan. Given a parent code, inspiration programs, and feedback from previous evaluations, the \emph{coder} generates a child while preserving the plan's ordered steps. It may revise concrete algorithms, data structures, parameters, solver settings, stopping rules, time allocations, and interactions between adjacent steps.
\end{itemize}

Detailed agent specifications are provided in Appendix~\ref{app:agents}; the prompt templates for initial code generation and plan-aware code evolution are given in Appendices~\ref{app:prompt-initial} and~\ref{app:prompt-evolution}, respectively.

\paragraph{Evaluation and interpreter feedback.}
Each child is evaluated on $\mathcal{D}$ under budget $B$ and checked for feasibility and execution failures. Based on these evaluation results, the Interpreter compares each validated child with its sampled parent in terms of objective quality, feasibility, runtime, material code changes, and step-level execution traces. Code-level feedback guides subsequent code updates on the selected island; plan-level feedback accumulates for the island's next plan review; and component-level evidence can be aggregated across islands. Separately, scheduling progress is measured against the island's best current-plan code rather than the sampled parent. Before the next update, the selected island's population, feedback, and scheduling statistics are updated. Retained programs and their reports enter the shared archive, whereas the active plans and populations of other islands remain unchanged. Appendix~\ref{app:interpreter} provides further details about the Interpreter, while Section~\ref{sec:evolution} describes how these updates are scheduled.

\subsection{Island-Based Evolution Procedure} \label{sec:evolution} 

\algo maintains $K$ islands, each associated with a current plan and an active population of validated programs. Following AdaEvolve \citep{cemri2026adaevolve}, we use an upper confidence bound (UCB) rule to allocate search across islands. Our controller additionally determines whether the selected island should continue improving code under its current plan or revise the plan itself. At initialization, each island is assigned a compositional plan together with a valid seed code (see Appendix~\ref{app:initialization}).

\paragraph{Island selection.}
Before each round, let $\bar r_k$ denote the discounted average reward for island $k$ and $n_k$ its cumulative number of completed code updates. Let $n_{\mathrm{sum}}=\sum_{k=1}^{K} n_k$ be the total count across islands. We select the next island as
$\label{eq:island_selection_main}
k^\star=
\arg\max_{k\in\{1,\ldots,K\}}
\left[
\bar r_k+
C\sqrt{\frac{\log n_{\mathrm{sum}}}{n_k}}
\right],
$ where the first term favors exploitation of islands that have recently earned higher rewards, while the second promotes exploration of islands that have been visited less often; $C$ controls the balance between the two. Initialization ensures $n_k>0$.
Our reward measures how much each code update helps an island close the gap to the current global best.  Any update that improves the island's best solution receives a positive reward, with larger rewards for improvements that close a larger fraction of its gap to the current global best. Non-improving updates receive zero reward. Exact reward and discounting definitions are provided in Appendix~\ref{app:evolution}.

\paragraph{Updating an island.}
We next describe the three mechanisms that govern evolution within and across islands: code refinement, plan review, and component evolution. The first two determine how the selected island is updated, while component evolution uses accumulated experience to refine the shared component pool.

After selecting an island, the controller decides whether to continue code-level improvement under the current plan or to review the plan itself. We measure recent progress by tracking whether successive code updates improve the island's current best code. If progress exceeds a plateau threshold $\tau$, the controller continues \textit{code refinement}; otherwise, it triggers a \textit{plan review}. The precise progress metric, smoothing rule, and reset behavior are described in Appendix~\ref{app:evolution}.

\textit{Code refinement.}
When the plan composition remains unchanged, evolution proceeds similarly to AdaEvolve \citep{cemri2026adaevolve}. Parents are drawn from the selected island's current population, while programs from other islands or earlier generations may serve as inspiration. The coder refines the implementation while preserving the plan's selected components and ordered subgoals.

\textit{Plan review.}
When progress stalls, \algo{} revisits the algorithmic composition. The planner uses accumulated feedback from the selected island, together with evidence from other islands, to propose a revised plan. Once validated, the revised plan initializes a new population for that island, followed by $W\geq1$ local code updates before global island selection resumes. Programs from the previous plan remain available as inspiration but are no longer eligible as parents.

\textit{Component evolution.}
Experience accumulated within the islands is also used to refine the shared component pool, allowing improved component descriptions and execution feedback to transfer across islands. These updates do not directly modify an island's current plan, which changes only through plan review. Algorithm~\ref{alg:evolution} in Appendix~\ref{app:evolution} presents the full evolution procedure.

\section{Experiments}

\subsection{Experimental Setup} \label{sec:exp_setup}

\paragraph{Dataset.}
We consider three families of data:
\textbf{1) Standard synthetic problems.} The 11 tasks are traveling salesman (TSP), clustered and uniform capacitated vehicle routing (C-CVRP and U-CVRP), orienteering (OP), job-shop scheduling (JSS), permutation flow-shop scheduling (PFSS), resource-constrained project scheduling (RCPS), offline bin packing (OBP), Max-Cut, maximum independent set (MIS), and decoupling-capacitor placement (Decap).
Instances use literature-based generation settings with task-specific adaptations; details are in Appendix~\ref{app:synthetic_tasks}.
\textbf{2) MIPLIB-derived problems.} Six complex real-world optimization problems are selected from MIPLIB-NL \cite{li2026constructing}: course scheduling (\texttt{comp12-2idx}, \texttt{comp21-2idx}), graph drawing (\texttt{graphdraw-grafo2}, \texttt{graphdraw-mainerd}), mining-project selection (\texttt{opm2-z12-s8}), and PCB assembly-line configuration (\texttt{sct32}).
All six were selected because Gurobi did not certify optimality within two hours in the initial screening.
Appendix~\ref{app:miplib_tasks} provides details.
\textbf{3) Nonlinear geometry problems.} We further evolve heuristics for four problems studied in AlphaEvolve \cite{novikov2025alphaevolve}: circle packing in a square, circle packing in a rectangle, min-max distance ratio, and hexagon packing (Appendix~\ref{app:nonlinear_datasets}).

\paragraph{Baselines.}
We compare against six LLM-based methods from two groups. The first comprises general-purpose program-evolution frameworks, including OpenEvolve \cite{openevolve}, AdaEvolve \cite{cemri2026adaevolve}, and EvoX \cite{liu2026evox}. The second comprises automatic heuristic design (AHD) methods, including EoH \cite{liu2024evolution}, ReEvo \cite{ye2024reevo}, and RefineEvo \cite{wu2026refineevo}. We instantiate the AHD baselines within problem-specific heuristic frameworks or solvers, including ant colony optimization (ACO) \cite{dorigo2006ant}, guided local search (GLS) \cite{arnold2019knowledge}, constructive heuristics, and Gurobi \cite{gurobi}.

\paragraph{Evaluation Metrics.}
All evolution runs, including baselines, use a budget of 200 generated programs and a two-minute evaluation limit per instance.
For method objective $f$ and Gurobi's 48-hour incumbent $f_{48\mathrm{h}}$, the signed gap is $g=s \cdot (f-f_{48\mathrm{h}})/|f_{48\mathrm{h}}|$, with $s=1$ for minimization and $s=-1$ for maximization. A positive gap indicates a worse objective than Gurobi's incumbent, while a negative gap indicates a better objective. For synthetic problems, we report the mean signed gap across five training instances sampled with different data seeds from the same task distribution; two validation instances select the program and four held-out test instances evaluate it.
MIPLIB is evaluated on five training instances per task; no validation or test evaluation is reported.
Evolution is guided by the fitness functions defined in Appendix~\ref{app:datasets}. For nonlinear problems, each heuristic evolves on an individual instance and is compared with the best objective reported in prior literature.
We use GPT-5.3-Codex for standard synthetic problems and GPT-5.6-sol for the more challenging MIPLIB-derived and nonlinear geometry problems.
Implementation details are in Appendix~\ref{app:implementation_details}.

\subsection{Main results}
\label{sec:main_results}
\label{sec:comparison_baselines}

\paragraph{Performance on Synthetic Dataset.}
Table~\ref{tab:synthetic_avg} summarizes the mean objective gaps across the 11 synthetic problems, comparing \algo{} with the code-evolution baselines OpenEvolve, AdaEvolve, and EvoX.
\algo{} achieves mean training and test gaps of $-2.22\%$ and $-1.45\%$, respectively; baseline training means range from $-0.15\%$ to $+1.22\%$, and their reported test means are all positive (OpenEvolve excludes OP from its test mean).
It discovers heuristics that produce better solutions than Gurobi does within 48 hours on 8 of the 11 training problem classes. When transferred to the test set, these heuristics remain effective, outperforming Gurobi's 48-hour solutions on 7 of the 11 problem classes.
As shown in Figure~\ref{fig:training_budget_curve}, \algo{} makes substantial gains early in evolution and maintains the lowest mean training gap throughout the subsequent search, indicating more efficient use of the program-generation budget.
Table~\ref{tab:synthetic_highlights} further examines five representative problem classes spanning routing, scheduling, and graph optimization.
\algo{} attains the best or tied-best training gaps on all five among the full-coverage program-evolution baselines, with clear test-time advantages on OP, C-CVRP, and RCPS, showing that these gains extend beyond the training instances.
Full results, evaluation coverage, and problem-selection details are provided in Appendix~\ref{app:synthetic_results}.

\makeatletter
\providecommand{\captionof}[1]{\def\@captype{#1}\caption}
\makeatother

\begin{table*}[!htbp]
  \centering
  \begin{minipage}[t]{0.42\textwidth}
    \vspace{0pt}
    \captionof{table}{Average synthetic performance. $^{\ddagger}$The OpenEvolve test mean excludes OP, whose reported test value has incomplete evaluation coverage.}
    \label{tab:synthetic_avg}
    \centering
    \setlength{\tabcolsep}{3pt}
    \renewcommand{\arraystretch}{1.15}
    \resizebox{\linewidth}{!}{\begin{tabular}{lrrrr}
      \toprule
      & \multicolumn{2}{c}{Mean gap (\%)} & \multicolumn{2}{c}{Beat Gurobi} \\
      \cmidrule(lr){2-3}\cmidrule(lr){4-5}
      Method & Train & Test & Train & Test \\
      \midrule
      OpenEvolve & $+0.71$ & $+0.41^{\ddagger}$ & $5/11$ & $4/11$ \\
      AdaEvolve & $-0.15$ & $+0.91$ & $7/11$ & $\mathbf{7/11}$ \\
      EvoX & $+1.22$ & $+2.24$ & $5/11$ & $4/11$ \\
      \algo{} & $\mathbf{-2.22}$ & $\mathbf{-1.45}$ & $\mathbf{8/11}$ & $\mathbf{7/11}$ \\
      \bottomrule
    \end{tabular}}
  \end{minipage}\hfill
  \begin{minipage}[t]{0.48\textwidth}
    \vspace{0pt}
    \centering
    \includegraphics[width=\linewidth]{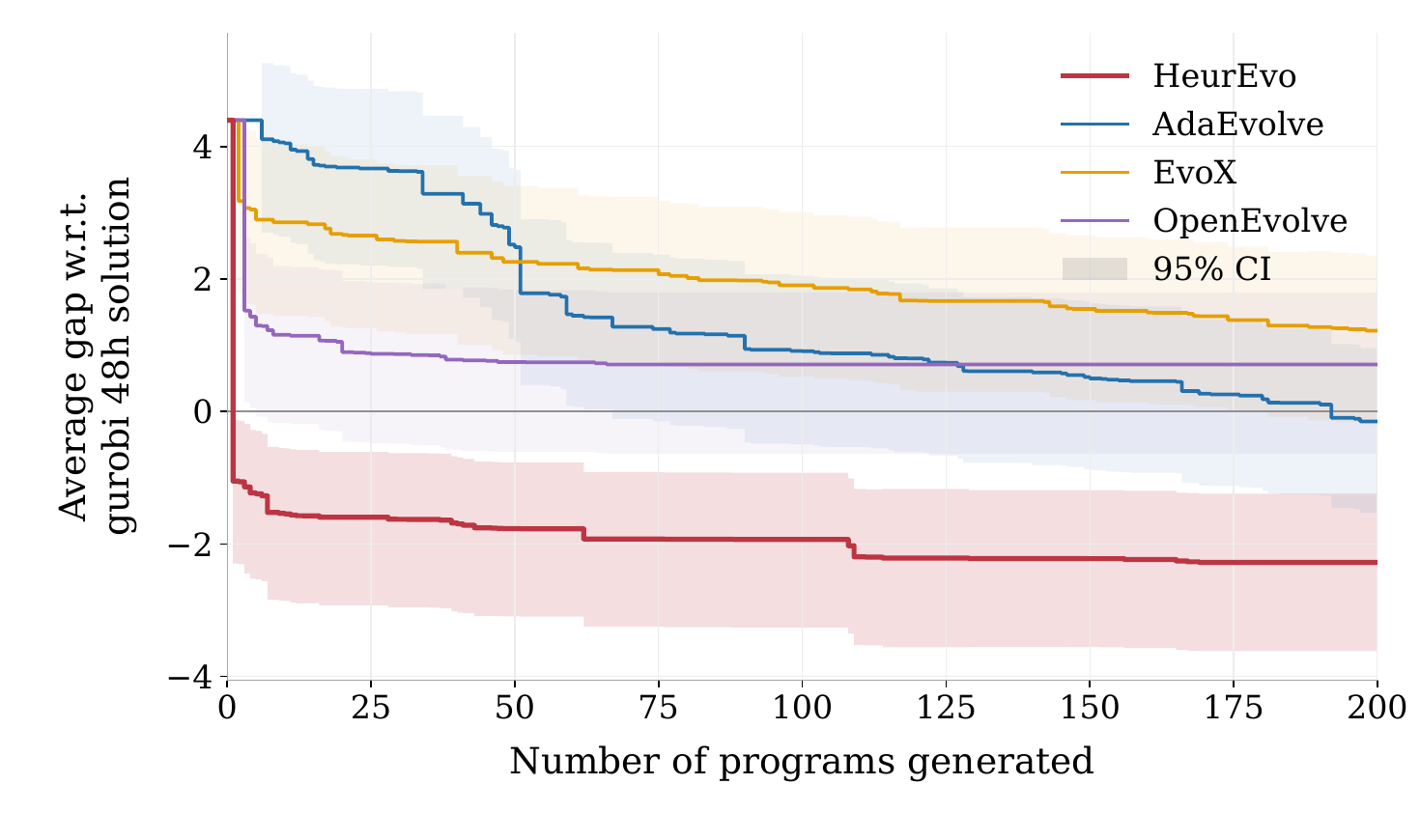}
    \vspace{-25pt}
    \captionof{figure}{Training gap versus program budget.}
    \label{fig:training_budget_curve}
  \end{minipage}
\end{table*}

\begin{table*}[!htbp]
  \centering
  \caption{Representative synthetic objective gaps (\%). Since AHD methods do not introduce a host function or an initial heuristic solution for the selected problems, their results are left blank. $^{\ddagger}$Incomplete evaluation coverage; excluded from the OpenEvolve test mean.}
  \label{tab:synthetic_highlights}
  \label{tab:main_results}
  \setlength{\tabcolsep}{3pt}
  \renewcommand{\arraystretch}{1.15}
  \resizebox{\textwidth}{!}{\begin{tabular}{l*{15}{r}}
      \toprule
      \multirow{2}{*}{Method}
      & \multicolumn{3}{c}{OP} & \multicolumn{3}{c}{C-CVRP}
      & \multicolumn{3}{c}{RCPS} & \multicolumn{3}{c}{PFSS}
      & \multicolumn{3}{c}{MIS} \\
      \cmidrule(lr){2-4}\cmidrule(lr){5-7}\cmidrule(lr){8-10}
      \cmidrule(lr){11-13}\cmidrule(lr){14-16}
      & Train & Val & Test & Train & Val & Test & Train & Val & Test
      & Train & Val & Test & Train & Val & Test \\
      \midrule
      EoH (GLS) & -- & -- & -- & -- & -- & -- & -- & -- & -- & \textbf{-2.65} & \textbf{-2.61} & -2.20 & -- & -- & -- \\
      ReEvo (ACO) & +4.16 & +1.32 & +7.40 & +6.58 & +9.77 & +8.28 & -- & -- & -- & -- & -- & -- & -- & -- & -- \\
      RefineEvo (ACO) & -- & -- & -- & +5.06 & +9.59 & +8.18 & -- & -- & -- & -- & -- & -- & -- & -- & -- \\
      \midrule
      \midrule
      
      OpenEvolve 
      & $+16.65$ & $+16.10$ & $+22.61^{\ddagger}$ 
      & $+1.93$ & $+4.57$ & $+2.96$ 
      & $+0.64$ & $\mathbf{-0.57}$ & $+1.74$ 
      & $-0.48$ & $+0.41$ & $-0.27$ 
      & $\mathbf{-0.04}$ & $+0.39$ & $\mathbf{+0.14}$ \\
      
      AdaEvolve 
      & $+16.92$ & $+15.73$ & $+21.75$ 
      & $-1.77$ & $-0.84$ & $-1.62$ 
      & $+1.49$ & $+1.21$ & $+0.70$ 
      & $-2.19$ & $-1.84$ & $\mathbf{-2.21}$ 
      & $+0.15$ & $+0.29$ & $+0.45$ \\
      
      EvoX 
      & $+17.96$ & $+16.94$ & $+21.57$ 
      & $+4.50$ & $+2.14$ & $+2.17$ 
      & $+2.70$ & $+3.05$ & $+0.91$ 
      & $-0.30$ & $-0.15$ & $-0.71$ 
      & $+0.47$ & $+1.17$ & $+1.88$ \\
      
      \algo{} 
      & $\mathbf{-3.80}$ & $\mathbf{-3.80}$ & $\mathbf{-1.94}$ 
      & $\mathbf{-3.36}$ & $\mathbf{-3.36}$ & $\mathbf{-4.38}$ 
      & $\mathbf{+0.12}$ & $+0.12$ & $\mathbf{0.00}$ 
      & $-2.25$ & $-2.25$ & $-1.85$ 
      & $\mathbf{-0.04}$ & $\mathbf{-0.04}$ & $\mathbf{+0.14}$ \\
      
      \midrule
      \bottomrule
    \end{tabular}}
  \par\smallskip
  \vspace{0pt}
\end{table*}

\paragraph{Performance on MIPLIB Problems.}
Table~\ref{tab:miplib_train_objectives} compares mean training objectives on the six MIPLIB-derived problems using GPT-5.6-sol. \algo{} achieves the better objective on four of the six problems. Relative to AdaEvolve, it reduces the objective by $10.70\%$ on \texttt{comp12-2idx}, $3.60\%$ on \texttt{graphdraw-grafo2}, and $4.71\%$ on \texttt{graphdraw-mainerd}, and increases it by $0.80\%$ on the maximization problem \texttt{opm2-z12-s8}. AdaEvolve achieves the better objective on \texttt{comp21-2idx} ($100.40$ versus $103.20$) and \texttt{sct32} ($-3.01$ versus $-2.77$). Appendix~\ref{app:miplib_objectives} reports the corresponding training trajectories.

\begin{table*}[!htbp]
  \centering
  \caption{Training objectives on MIPLIB.}
  \label{tab:miplib_train_objectives}
  \label{tab:miplib_results}
  \setlength{\tabcolsep}{3pt}
  \renewcommand{\arraystretch}{1.2}
  \resizebox{0.8\linewidth}{!}{\begin{tabular}{@{}l*{6}{r}@{}}
      \toprule
      Method
      & \shortstack{\texttt{comp12-2idx}\\$\downarrow$}
      & \shortstack{\texttt{comp21-2idx}\\$\downarrow$}
      & \shortstack{\texttt{graphdraw-}\\\texttt{grafo2} $\downarrow$}
      & \shortstack{\texttt{graphdraw-}\\\texttt{mainerd} $\downarrow$}
      & \shortstack{\texttt{opm2-z12-s8}\\$\uparrow$}
      & \shortstack{\texttt{sct32}\\$\downarrow$} \\
      \midrule
      AdaEvolve & 366.20 & \textbf{100.40} & 85061.30 & 47940.40 & 58057.20 & \textbf{-3.01} \\
      \algo{} & \textbf{327.00} & 103.20 & \textbf{82003.10} & \textbf{45681.00} & \textbf{58519.40} & -2.77 \\
      \bottomrule
    \end{tabular}}
  \vspace{0pt}
\end{table*}

\paragraph{Performance on Nonlinear Problems.}

Table~\ref{tab:nonlinear_training} reports eight selected nonlinear task-size configurations. The reported results use a \algo{} phase followed by AdaEvolve refinement. They improve on the standalone AdaEvolve baseline in all eight configurations and are numerically better than the listed reference values in five: rectangle packing with $n=26$, the three distance-ratio configurations, and hexagon packing with $n=12$. Rectangle packing with $n=21$ matches the reference at the displayed precision, while the hexagon-packing objectives for $n=11$ and $n=14$ remain slightly worse than their references. Table~\ref{tab:nonlinear_full_train} reports all 14 configurations and the reference sources. Comparisons against rounded references are limited to the reported precision and do not by themselves establish an improvement over an unrounded solution.

\begin{table*}[!htbp]
\centering
\vspace{-10pt}
\caption{Nonlinear objectives versus reported best values.
\underline{Underline} indicates the part better than AdaEvolve;
\textbf{Bold} indicates the part better than the reported best-known value.}
\vspace{-5pt}
\label{tab:nonlinear_training}
\label{fig:nonlinear_training}
\setlength{\tabcolsep}{2pt}
\renewcommand{\arraystretch}{1.2}
\resizebox{\linewidth}{!}{\begin{tabular}{@{}l*{8}{r}@{}}
\toprule
\multirow{2}{*}{Method}
& \multicolumn{2}{c}{Circle packing-rectangle $\uparrow$}
& \multicolumn{3}{c}{Min-max distance ratio $\downarrow$}
& \multicolumn{3}{c}{Hexagon packing $\downarrow$} \\
\cmidrule(lr){2-3}\cmidrule(lr){4-6}\cmidrule(lr){7-9}
& $n=21$ & $n=26$
& $n=16,d=2$ & $n=21,d=2$ & $n=22,d=2$
& $n=11$ & $n=12$ & $n=14$ \\
\midrule
SOTA
& 2.3658323759
& 2.6393205590
& 12.8892299\phantom{000}
& 17.77499\phantom{00000}
& 19.05398\phantom{00000}
& 3.9245008973
& 3.9416420675
& 4.2689949573 \\
AdaEvolve
& 2.3642481267
& 2.6355313151
& 12.8892299077
& 17.7749805542
& 19.0539845098
& 3.9326758559
& 3.9416423004
& 4.2723920772 \\
\algo{}
& 2.36\underline{58323759}
& 2.63\underline{93205\textbf{643}}
& 12.889229\textbf{\underline{1072}}
& 17.7749\textbf{80\underline{3417}}
& 19.0539\textbf{\underline{769136}}
& 3.9\underline{245011309}
& 3.941642\underline{06\textbf{25}}
& 4.2\underline{689950806} \\
\bottomrule
\end{tabular}}
\vspace{-10pt}
\end{table*}

\subsection{Ablation Studies}

\paragraph{Effectiveness of Each Module.}
To assess the contribution of each evolutionary component in HeurEvo, we compare the following five ablation variants:
1)~\emph{Fixed plan:} no plan or component evolution;
2)~\emph{Component only:} evolve components while keeping plans fixed;
3)~\emph{Always-plan:} review plans every round with fixed components;
4)~\emph{Always-plan + component:} combine per-round plan reviews with component evolution;
5)~\emph{Adaptive plan only:} review plans when progress stalls, with fixed components.
Always-plan variants review the selected island's plan every round; adaptive review occurs only when its smoothed relative improvement falls below the plateau threshold, otherwise continuing code refinement.
The full pipeline, \algo{} (Ours), combines adaptive plan review with component evolution.
As shown in Table~\ref{tab:component_ablation}, the full pipeline achieves the best mean training, validation, and test gaps of $-2.22\%$, $-2.22\%$, and $-1.45\%$, respectively.
Compared with the Adaptive plan only, it lowers the training and test gaps by $0.87$ and $1.24$ percentage points, respectively; its test gap is also $0.75$ percentage points lower than the Always-plan + component.
These results support combining adaptive plan review with component evolution for both training performance and generalization.

\begin{table}[t]
\centering
\caption{Component ablation of \algo{}.}
\label{tab:component_ablation}
\setlength{\tabcolsep}{2.7pt}
\renewcommand{\arraystretch}{1.18}
\resizebox{\textwidth}{!}{\begin{tabular}{@{}l*{15}{r}@{}}
\toprule
\multirow{2}{*}{Variant} &
\multicolumn{3}{c}{Averaged performance} &
\multicolumn{3}{c}{OP} &
\multicolumn{3}{c}{C-CVRP} &
\multicolumn{3}{c}{RCPS} &
\multicolumn{3}{c}{PFSS} \\
\cmidrule(lr){2-4}\cmidrule(lr){5-7}\cmidrule(lr){8-10}\cmidrule(lr){11-13}\cmidrule(lr){14-16}
& Train & Val & Test & Train & Val & Test & Train & Val & Test & Train & Val & Test & Train & Val & Test \\
\midrule
Fixed plan & $-1.19$ & $+0.06$ & $+0.25$ & $+1.94$ & $+8.1$ & $+7.97$ & $-3.15$ & $-2.3$ & $-3.26$ & $+1.39$ & $+1.8$ & $+0.39$ & $-0.83$ & $-1.0$ & $-0.50$ \\
Component only & $-1.45$ & $-0.38$ & $+0.08$ & $-1.67$ & $+0.9$ & $+3.86$ & $-3.09$ & $-2.4$ & $-3.47$ & $+1.37$ & $+1.2$ & $+1.61$ & $-0.79$ & $-0.3$ & $-0.69$ \\

Always-plan & $-1.67$ & $-0.36$ & $-0.27$ & $+0.45$ & $+8.0$ & $+8.12$ & $-2.84$ & $-2.0$ & $-2.95$ & $+0.50$ & $\mathbf{0.0}$ & $+0.43$ & $-1.52$ & $-1.6$ & $-1.67$ \\
Always-plan + component & $-1.89$ & $-1.38$ & $-0.70$ & $-2.67$ & $\mathbf{-8.9}$ & $\mathbf{-2.39}$ & $-3.20$ & $-2.6$ & $-3.62$ & $\mathbf{-0.35}$ & $\mathbf{0.0}$ & $+0.43$ & $-1.16$ & $-0.9$ & $-0.93$ \\

Adaptive plan only & $-1.35$ & $-1.01$ & $-0.21$ & $-0.87$ & $-3.8$ & $+3.76$ & $-2.48$ & $-2.2$ & $-3.00$ & $+0.87$ & $+0.1$ & $+0.65$ & $-1.48$ & $-1.0$ & $-0.64$ \\

\midrule
\algo{} (Ours) & $\mathbf{-2.22}$ & $\mathbf{-2.22}$ & $\mathbf{-1.45}$ & $\mathbf{-3.80}$ & $-3.80$ & $-1.94$ & $\mathbf{-3.36}$ & $\mathbf{-3.36}$ & $\mathbf{-4.38}$ & $+0.12$ & $+0.12$ & $\mathbf{0.00}$ & $\mathbf{-2.25}$ & $\mathbf{-2.25}$ & $\mathbf{-1.85}$ \\

\bottomrule
\end{tabular}}
\vspace{0pt}
\end{table}

\begin{wraptable}{r}{0.4\textwidth}
    \centering
    \vspace{0pt}
    \caption{Island initialization ablation.}
    \vspace{-5pt}
    \label{tab:island_initialization}
    \footnotesize
    \setlength{\tabcolsep}{2.5pt}
    \renewcommand{\arraystretch}{1.05}
    \resizebox{\linewidth}{!}{
    \begin{tabular}{@{}lrrr@{}}
        \toprule
        \multirow{2}{*}{Variant}
        & \multicolumn{3}{c}{Average gap (\%)} \\
        \cline{2-4} \addlinespace[3.5pt]
        & Train & Val & Test \\
        \midrule
        Single island & $-1.31$ & $-0.98$ & $-0.75$ \\
        Shared plan & $-1.76$ & $-1.24$ & $-1.23$ \\
        Partitioned library & $-1.41$ & $-1.69$ & $\mathbf{-1.57}$ \\
        \midrule
        \algo{} (Ours)
          & $\mathbf{-2.22}$ & $\mathbf{-2.22}$ & $-1.45$ \\
        \bottomrule
    \end{tabular}
    }
\end{wraptable}

\paragraph{Effect of Island Count and Plan Initialization.}
Table~\ref{tab:island_initialization} compares three alternatives:
1)~\emph{Single island:} one plan with the full component library;
2)~\emph{Shared plan:} five islands initialized with the same plan and full library;
3)~\emph{Partitioned library:} five distinct plans generated from separate component-library partitions.
Our default uses five distinct initial plans generated from the full library.
It achieves the lowest training gap ($-2.22\%$), improving on these alternatives by $0.91$, $0.46$, and $0.81$ percentage points, respectively, supporting both initial-plan diversity and access to the full component library.

We include the full table of ablation experiments in Appendix~\ref{app:synthetic_ablations}.

\section{Conclusion}

We presented \algo{}, a plan--code--component co-evolution framework for designing hybrid, solver-augmented heuristics for time-critical optimization. \algo{} jointly evolves algorithmic structure, implementation, and reusable components through execution feedback, using plan-conditioned islands, adaptive plan review, and a shared component library. With a two-minute execution budget, the evolved heuristics outperform Gurobi's 48-hour solutions on seven of eleven synthetic tasks and several competitive LLM-based evolution baselines, with additional gains on MIPLIB and nonlinear problems. These results demonstrate the value of jointly evolving algorithmic structure and implementation rather than optimizing heuristic components in isolation.

\section*{AI Use Statement}

We used large language models (LLMs) in the proposed framework and experiments, to polish the authors' initial manuscript draft, and to generate code for constructing optimization instances. The authors reviewed the revised text, and the generated instances were checked against the experimental assumptions and requirements using human-written evaluators. Coding agents, including Codex and Claude Code, assisted with implementation, which human experts reviewed for consistency with the described framework. The authors take full responsibility for the manuscript and research artifacts, including the accuracy of claims and results, the validity of data and code, and the correctness and attribution of references.

\section*{Reproducibility Statement}

To support reproducibility, we document the experimental setup, models, and implementation details in Section \ref{sec:exp_setup} and Appendix \ref{app:additional_experiments}, with experimental hyperparameters summarized in Table~\ref{tab:hyperparameters}. Appendix \ref{app:datasets} describes the data sources, generation and synthesis procedures, and dataset splits, while Appendix \ref{app:prompts} provides the prompt templates. We will publicly release the code and datasets, including the data-generation scripts.

\bibliography{references}
\bibliographystyle{references}

\newpage
\appendix

\renewcommand{\thetable}{\Alph{section}\arabic{table}}
\renewcommand{\theHtable}{appendix.\Alph{section}.\arabic{table}}

\section{Additional Experiments}
\label{app:additional_experiments}

\paragraph{Implementation Details.}
\label{app:implementation_details}
We use GPT-5.3-Codex \cite{openai2026gpt53codex} for the synthetic experiments and GPT-5.6-sol \cite{openai2026gpt56sol} for both the MIPLIB-NL and nonlinear experiments. All experiments are conducted on a multi-node computing cluster, with each node equipped with 128 CPU cores and 1000 GB of memory. During evolution, each generated program is executed single-threaded on a single CPU core and is subject to a 120-second runtime limit per problem instance. 
Gurobi 13.0.2 is the only optimization solver permitted in the generated programs. All evolution experiments, including the baselines, are limited to generating 200 programs. For the synthetic experiments, generated programs are evaluated on the validation set during model selection. After evolution, the program with the best validation performance is selected and evaluated once on the held-out test set. Table~\ref{tab:hyperparameters} summarizes the complete hyperparameter configuration used for \algo{}. 

\begin{table}[!htbp]
\centering
\small
\setlength{\tabcolsep}{5pt}
\renewcommand{\arraystretch}{1.12}
\caption{Hyperparameters of HeurEvo. Symbols follow Appendix \ref{app:method-details}; a dash marks a quantity the analysis does not name. Values describe the default configuration; the ablation studies vary the specified components.}
\label{tab:hyperparameters}
\begin{tabular}{@{}l l l r@{}}
\toprule
& \textbf{Symbol} & \textbf{Description} & \textbf{Value} \\
\midrule
\multicolumn{4}{@{}l}{\textit{Search budget and initialization}} \\
& $K$              & Number of islands (one initial plan per island)        & 5 \\
& --               & Validated seed programs generated per plan              & 1 \\
& $M_{\mathrm{dbg}}$ & Debugging attempts per generation call                & 3 \\
& $B$              & Per-instance execution budget (seconds)                 & 120 \\
\midrule
\multicolumn{4}{@{}l}{\textit{Island scheduling (Eqs.~C1, C3)}} \\
& $\rho$           & Discount factor of $R$, $V$, and $H$                    & 0.9 \\
& $C$              & UCB exploration coefficient                             & 1.41 \\
& $W$              & Local code visits per island at initialization (Alg.~C1) & 3 \\
\midrule
\multicolumn{4}{@{}l}{\textit{Code evolution (OpenEvolve backend)}} \\
& $|\mathcal{I}^{(k)}_{j,h}|$ & Inspiration programs sampled per visit       & 2 \\
& --               & Top-ranked programs shown in the prompt                 & 2 \\
& --               & Ancestor programs shown in the prompt (metrics only)    & 3 \\
& --               & Elite / exploration / exploitation sampling ratio       & 0.1 / 0.2 / 0.7 \\
& --               & MAP-Elites feature dimensions                           & complexity, diversity \\
& --               & MAP-Elites bins per dimension                           & 10 \\
& --               & Population size / archive size                          & 1000 / 100 \\
\midrule
\multicolumn{4}{@{}l}{\textit{Plan evolution (Alg.~C2)}} \\
& $\tau$           & Plateau threshold; $H^{(k)}_{j-1}<\tau$ triggers a plan review & 0.01 \\
& $W$              & Local code visits after an accepted plan proposal       & 3 \\
& --               & Evidence: top programs under the current plan           & 2 \\
& --               & Evidence: inspiration programs from the same island     & 3 \\
& --               & Evidence: top programs from each other island           & 0 \\
& --               & Evidence: inspiration programs across islands           & 5 \\
\midrule
\multicolumn{4}{@{}l}{\textit{Component evolution (Eq.~C8)}} \\
& --               & Rounds between Component Improver calls                 & 1 \\
& $C_I$            & Sensitivity to the contribution score $Q_j(s)$          & 2.0 \\
& $I_{\min}$       & Minimum review probability                              & 0.1 \\
& $I_{\max}$       & Maximum review probability                              & 0.7 \\
\bottomrule
\end{tabular}
\end{table}

\subsection{Standard Synthetic problems}
\label{app:synthetic_results}

Tables~\ref{tab:synthetic_full_train} and~\ref{tab:synthetic_full_test} report all 11 problems on the training and test sets, respectively. EoH \cite{liu2024evolution}, ReEvo \cite{ye2024reevo}, and RefineEvo \cite{wu2026refineevo} retain separate constructive, ACO, and GLS variants wherever supported. In other words, their sparse task coverage prevents them from running all 11 tasks and excludes them from Table~\ref{tab:synthetic_avg}, which reports the mean objective gap across the problem classes (excluding OP from the OpenEvolve test mean because of incomplete evaluation coverage). Among the full-coverage methods, \algo{} achieves the lowest mean gap on both splits, with $-2.22\%$ on training and $-1.45\%$ on test, compared with $-0.15\%$ and $+0.91\%$ for AdaEvolve, respectively. This corresponds to reductions of $2.07$ and $2.36$ percentage points over AdaEvolve. At the individual-problem level, \algo{} obtains the best or tied-best gap on six of the 11 problems on both training and test: C-CVRP, U-CVRP, OP, Decap, RCPS, and MIS. It also achieves negative gaps on eight training problems and seven test problems, showing that the gains are not confined to the training instances. The largest improvement over the full-coverage baselines occurs on OP, where \algo{} changes the gap from $+16.92\%$ for AdaEvolve to $-3.80\%$ on training and from $+21.75\%$ to $-1.94\%$ on test.

\begin{table*}[!htbp]
  \centering
  \caption{Full synthetic training gaps (\%).}
  \label{tab:synthetic_full_train}
  \setlength{\tabcolsep}{3pt}
  \renewcommand{\arraystretch}{1.12}
  \resizebox{\textwidth}{!}{\begin{tabular}{l*{11}{r}}
      \toprule
      Method & C-CVRP & U-CVRP & TSP & BP & OP & Decap & JSS & PFSS & RCPS & Max-Cut & MIS \\
      \midrule
      EoH (Constr.) & -- & -- & -- & +3.53 & -- & -- & -- & -- & -- & -- & -- \\
      EoH (GLS) & -- & -- & +0.68 & -- & -- & -- & -- & \textbf{-2.65} & -- & -- & -- \\
      ReEvo (ACO) & +6.58 & +9.66 & +8.60 & +1.14 & +4.16 & -- & -- & -- & -- & -- & -- \\
      ReEvo (Constr.) & -- & -- & +16.02 & -- & -- & -- & -- & -- & -- & -- & -- \\
      ReEvo (GLS) & -- & -- & \textbf{-0.83} & -- & -- & -- & -- & -- & -- & -- & -- \\
      RefineEvo (ACO) & +5.06 & +9.10 & +9.60 & -- & -- & -- & -- & -- & -- & -- & -- \\
      RefineEvo (Constr.) & +12.13 & +16.10 & +20.07 & +2.29 & -- & -- & -- & -- & -- & -- & -- \\
      RefineEvo (GLS) & -- & -- & \textbf{-0.83} & -- & -- & -- & -- & -- & -- & -- & -- \\
      OpenEvolve & +1.93 & -1.16 & +1.53 & +0.45 & +16.65 & +1.63 & \textbf{-12.54} & -0.48 & +0.64 & -0.80 & \textbf{-0.04} \\
      AdaEvolve & -1.77 & -2.06 & -0.80 & \textbf{-0.45} & +16.92 & +1.63 & -12.46 & -2.19 & +1.49 & \textbf{-2.08} & +0.15 \\
      EvoX & +4.50 & -1.63 & +1.00 & -0.20 & +17.96 & +1.50 & -12.40 & -0.30 & +2.70 & -0.17 & +0.47 \\
      \algo{} & \textbf{-3.36} & \textbf{-3.19} & -0.82 & +0.45 & \textbf{-3.80} & \textbf{+1.40} & -11.60 & -2.25 & \textbf{+0.12} & -1.32 & \textbf{-0.04} \\
      \bottomrule
    \end{tabular}}
\end{table*}

\begin{table*}[!htbp]
  \centering
  \caption{Full synthetic test gaps (\%).}
  \label{tab:synthetic_full_test}
  \setlength{\tabcolsep}{3pt}
  \renewcommand{\arraystretch}{1.12}
  \resizebox{\textwidth}{!}{\begin{tabular}{l*{11}{r}}
      \toprule
      Method & C-CVRP & U-CVRP & TSP & BP & OP & Decap & JSS & PFSS & RCPS & Max-Cut & MIS \\
      \midrule
      EoH (Constr.) & -- & -- & -- & +3.30 & -- & -- & -- & -- & -- & -- & -- \\
      EoH (GLS) & -- & -- & +0.60 & -- & -- & -- & -- & -2.20 & -- & -- & -- \\
      ReEvo (ACO) & +8.28 & +13.30 & +9.96 & +1.56 & +7.40 & -- & -- & -- & -- & -- & -- \\
      ReEvo (Constr.) & -- & -- & +15.37 & -- & -- & -- & -- & -- & -- & -- & -- \\
      ReEvo (GLS) & -- & -- & \textbf{-0.79} & -- & -- & -- & -- & -- & -- & -- & -- \\
      RefineEvo (ACO) & +8.18 & +12.26 & +12.72 & -- & -- & -- & -- & -- & -- & -- & -- \\
      RefineEvo (Constr.) & +9.91 & +19.12 & +23.22 & +2.37 & -- & -- & -- & -- & -- & -- & -- \\
      RefineEvo (GLS) & -- & -- & \textbf{-0.79} & -- & -- & -- & -- & -- & -- & -- & -- \\
      OpenEvolve & +2.96 & -1.30 & +4.67 & +0.69 & +22.61 & +2.04 & -6.30 & -0.27 & +1.74 & -0.31 & \textbf{+0.14} \\
      AdaEvolve & -1.62 & -2.18 & -0.38 & \textbf{-0.32} & +21.75 & +2.02 & \textbf{-6.42} & \textbf{-2.21} & +0.70 & \textbf{-1.78} & +0.45 \\
      EvoX & +2.17 & -0.20 & +3.14 & -0.06 & +21.57 & +1.95 & -6.16 & -0.71 & +0.91 & +0.11 & +1.88 \\
      \algo{} & \textbf{-4.38} & \textbf{-3.48} & -0.70 & +0.69 & \textbf{-1.94} & \textbf{+1.90} & -5.52 & -1.85 & \textbf{0.00} & -0.85 & \textbf{+0.14} \\
      \bottomrule
    \end{tabular}}
\end{table*}

\subsection{MIPLIB-derived problems}
\label{app:miplib_objectives}

\begin{figure*}[t]
  \centering
  \subfloat[comp12-2idx]{\includegraphics[width=0.32\textwidth]{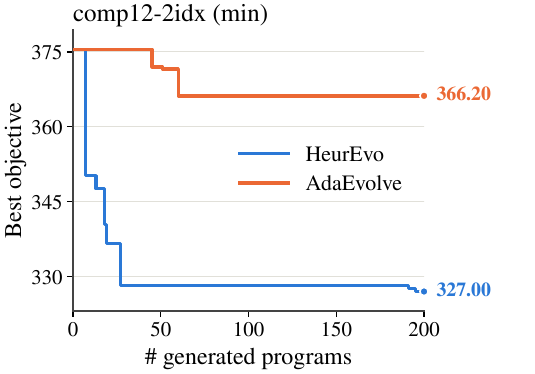}}
  \hfill
  \subfloat[comp21-2idx]{\includegraphics[width=0.32\textwidth]{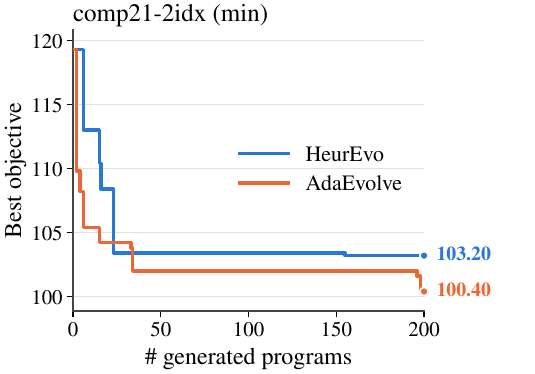}}
  \hfill
  \subfloat[graphdraw-grafo2]{\includegraphics[width=0.32\textwidth]{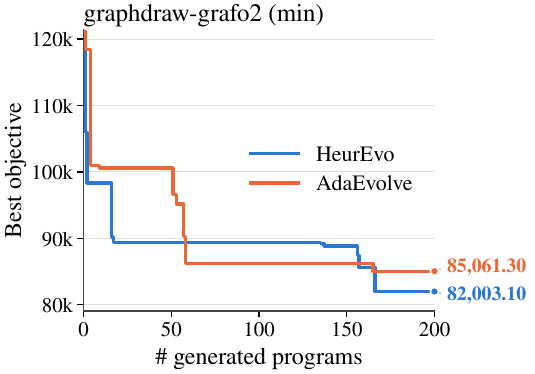}}
  \\[0.5em]
  \subfloat[graphdraw-mainerd]{\includegraphics[width=0.32\textwidth]{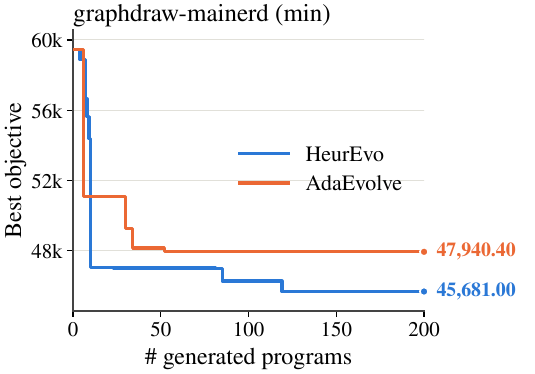}}
  \hfill
  \subfloat[opm2-z12-s8]{\includegraphics[width=0.32\textwidth]{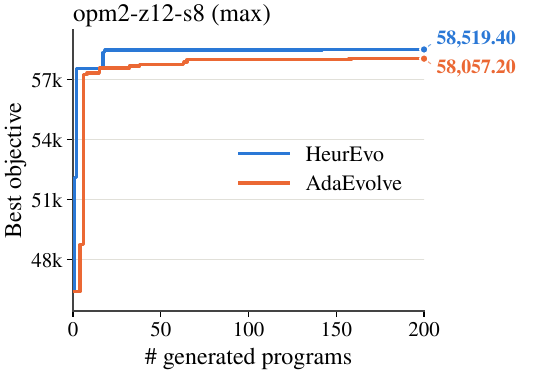}}
  \hfill
  \subfloat[sct32]{\includegraphics[width=0.32\textwidth]{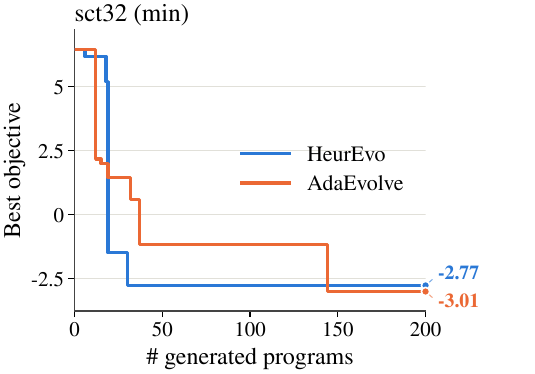}}
  \caption{Best objective versus number of generated programs for HeurEvo and AdaEvolve
    (solver model GPT-5.6) on the six MIPLIB instances. Objectives are averaged over the
    five training seeds; the value at the end of each curve is the final best objective.}
  \label{fig:gpt56-curves}
\end{figure*}

Earlier in Table~\ref{tab:miplib_train_objectives}, we report the averaged training objectives on MIPLIB problems. Figure~\ref{fig:gpt56-curves} demonstrates the training performance of each method, i.e., mean objectives vs. the number of generated programs. The trajectories reveal different improvement patterns across problems. On \texttt{comp12-2idx}, \texttt{graphdraw-mainerd}, and \texttt{opm2-z12-s8}, \algo{} establishes an advantage within the first few dozen generated programs and maintains it through the end of the run. On \texttt{graphdraw-grafo2}, AdaEvolve leads during the middle of training, but late-stage improvements allow \algo{} to finish with a lower objective of $82003.10$, compared with $85061.30$ for AdaEvolve. However, \algo{} does not dominate on every problem. AdaEvolve achieves a lower final objective on \texttt{comp21-2idx}, reaching $100.40$ compared with $103.20$ for \algo{}. On \texttt{sct32}, \algo{} reaches an objective near $-2.77$ within roughly 30 generated programs but makes little subsequent progress, whereas AdaEvolve overtakes it after approximately 140 programs and finishes at $-3.01$. These contrasting trajectories show that early advantages do not always determine the final ranking and that improvements after extended plateaus can remain important.

\subsection{Non-linear Geometry Problems}
\label{app:nonlinear_results}

The reported nonlinear results combine a \algo{} phase with subsequent AdaEvolve refinement. Table~\ref{tab:nonlinear_full_train} contains all 14 nonlinear task-size configurations under GPT-5.6-sol, which is extended from the main Table~\ref{tab:nonlinear_training}. In this table, we offer rich information; for example, we show the output of each best-known solution. Moreover, we demonstrate the performance where the proposed \algo achieves poorer performance. Across all 14 configurations, \algo{} achieves better objectives than AdaEvolve in 11, matches it at the reported precision in one, and performs worse in two. Its improvements cover all three rectangle-packing configurations, all four distance-ratio configurations, and four of the five hexagon-packing configurations. However, these improvements vary in magnitude. For example, on hexagon packing with $n=11$, \algo{} reduces the AdaEvolve objective to $3.9245011309$ at the second decimal place, approaching the best-known solution. In contrast, its improvement over AdaEvolve for $n=16$ is only $1.812 \times 10^{-7}$, and the resulting objective of $4.5324919358$ remains slightly above the reported reference. 

\begin{table*}[!htbp]
  \centering
  \caption{Complete nonlinear training objectives. For \algo{}, \underline{Underline} indicates the part better than AdaEvolve, and \textbf{Bold} indicates the part better than the reported best-known value.}
  \label{tab:nonlinear_full_train}
  \setlength{\tabcolsep}{4pt}
  \renewcommand{\arraystretch}{1.15}
  \resizebox{\textwidth}{!}{\begin{tabular}{lcclrrr}
      \toprule
      Problem & $n$ & $d$ & Best result source & Reported reference & AdaEvolve & \algo{} \\
      \midrule
      Circle packing (rectangle) $\uparrow$ & 21 & -- &
      EinsteinArena$^{a}$ &
      2.3658323759185156 & 2.3642481267 & 2.36\underline{58323759} \\
      Circle packing (rectangle) $\uparrow$ & 26 & -- &
      Lai$^{b}$ &
      2.639320558987759336 & 2.6355313151 & 2.63\underline{93205\textbf{643}} \\
      Circle packing (rectangle) $\uparrow$ & 27 & -- &
      Dutton / Lai$^{b,c}$ &
      2.691523360671018652 & 2.6834410277 & 2.68\underline{91174195} \\
      Circle packing (square) $\uparrow$ & 26 & -- &
      Packomania$^{d}$ &
      2.635983084919 & 2.6359830484 & 2.6359775049 \\
      Circle packing (square) $\uparrow$ & 32 & -- &
      \citet{berthold2026geometry}$^{d}$ &
      2.939572771205 & 2.9395727706 & 2.9395727706 \\
      Distance ratio $\downarrow$ & 14 & 3 &
      Sun--Samanta$^{e}$ &
      4.165 & 4.1657834746 & 4.16578\underline{20926} \\
      Distance ratio $\downarrow$ & 16 & 2 &
      Together-AI$^{f}$ &
      12.8892299 & 12.8892299077 & 12.889229\underline{\textbf{1072}} \\
      Distance ratio $\downarrow$ & 21 & 2 &
      \citet{berthold2026geometry} &
      17.77499 & 17.7749805542 & 17.7749\textbf{80\underline{3417}} \\
      Distance ratio $\downarrow$ & 22 & 2 &
      \citet{berthold2026geometry} &
      19.05398 & 19.0539845098 & 19.0539\underline{\textbf{769136}} \\
      Hexagon packing $\downarrow$ & 11 & -- &
      Schellhorn$^{g}$ &
      3.9245008972987525 & 3.9326758559 & 3.9\underline{245011309} \\
      Hexagon packing $\downarrow$ & 12 & -- &
      \citet{berthold2026packing}$^{h}$ &
      3.941642067536949 & 3.9416423004 & 3.941642\underline{06\textbf{25}} \\
      Hexagon packing $\downarrow$ & 14 & -- &
      \citet{berthold2026packing}$^{h}$ &
      4.268994957252001 & 4.2723920772 & 4.2\underline{689950806} \\
      Hexagon packing $\downarrow$ & 15 & -- &
      \citet{kravatskiy2026improvevolve}$^{g,h}$ &
      4.447271148987105 & 4.4472718765 & 4.4494139689 \\
      Hexagon packing $\downarrow$ & 16 & -- &
      \citet{kravatskiy2026improvevolve}$^{g}$ &
      4.52746 & 4.5324921170 & 4.53249\underline{19358} \\
      \bottomrule
    \end{tabular}}
  \par\smallskip
  \begin{minipage}{\textwidth}
    \scriptsize
    \raggedright
    $^{a}$EinsteinArena solution and verification notebook:
    \url{https://github.com/togethercomputer/EinsteinArena-new-SOTA/blob/c388c6f7408c886311940896713339a1a70c2394/circles-rectangle/analysis.ipynb}.\par
    $^{b}$Lai's public rectangle configurations (files \texttt{coords\_n26.txt} and \texttt{coords\_n27.txt}):
    \url{https://github.com/lshhhhhhh/circle-packing-records/tree/8dfa6449385015a58aabfbf84f9dd5f063bfa96b}.\par
    $^{c}$Friedman's rectangle record list (including Dutton's $n=27$ record):
    \url{https://erich-friedman.github.io/packing/cirRrec/}.\par
    $^{d}$Packomania's circle-in-square numerical records:
    \url{https://www.packomania.com/csqv/csqv.html}.\par
    $^{e}$Friedman's three-dimensional distance-ratio records:
    \url{https://erich-friedman.github.io/packing/maxmin3/}.\par
    $^{f}$Together-AI's reported score on the EinsteinArena leaderboard
    for the two-dimensional distance-ratio problem ($n=16$):
    \url{https://einsteinarena.com/problems/min-distance-ratio-2d}.\par
    $^{g}$Friedman's hexagon-in-hexagon records:
    \url{https://erich-friedman.github.io/packing/hexinhex/}.\par
    $^{h}$Public numerical configurations accompanying the packing paper
    (subdirectories \texttt{n12}, \texttt{n14}, and \texttt{n15}):
    \url{https://github.com/DominikKamp/Packing/tree/27f71632d8655ae4b6a049410a401976748b94e3/polygon/l6/m6}.
  \end{minipage}
\end{table*}

\subsection{Full ablation results on synthetic problems}
\label{app:synthetic_ablations}

Earlier in Table~\ref{tab:component_ablation}, we show the effectiveness of various components by showing the mean objective gap across all 11 synthetic tasks and choose the four most representative problems to analyze. Tables~\ref{tab:ablation_full_train} and~\ref{tab:ablation_full_test} extend from Table~\ref{tab:component_ablation} with all 11 synthetic tasks performance. Across all 11 tasks, \algo{} achieves the lowest gap on five tasks in each split, leading on C-CVRP, U-CVRP, TSP, and PFSS in both training and testing. It outperforms \emph{Adaptive plan only} on nine tasks and \emph{Component only} on eight tasks in each split. Compared with \emph{Always-plan + component}, it achieves lower gaps on seven training tasks and six test tasks. These results support combining adaptive plan review with component evolution for both training performance and generalization. 

\begin{table*}[!htbp]
  \centering
  \caption{Synthetic ablation training gaps (\%).}
  \label{tab:ablation_full_train}
  \setlength{\tabcolsep}{3pt}
  \renewcommand{\arraystretch}{1.12}
  \resizebox{\textwidth}{!}{\begin{tabular}{l*{11}{r}}
      \toprule
      Method & C-CVRP & U-CVRP & TSP & BP & OP & Decap & JSS & PFSS & RCPS & Max-Cut & MIS \\
      \midrule
      Fixed plan & -3.15 & -2.25 & +1.23 & +0.45 & +1.94 & +1.47 & -12.29 & -0.83 & +1.39 & -1.19 & +0.19 \\
      Adaptive plan only & -2.48 & -0.68 & +1.11 & +0.45 & -0.87 & +1.43 & \textbf{-12.32} & -1.48 & +0.87 & -1.18 & +0.27 \\
      Component only & -3.09 & -2.47 & +1.98 & +0.40 & -1.67 & \textbf{+1.30} & -12.28 & -0.79 & +1.37 & -1.08 & +0.35 \\
      Earlier full configuration & -3.04 & -1.82 & +1.45 & +0.45 & -0.75 & +1.45 & \textbf{-12.32} & -1.15 & +0.82 & -1.83 & +0.23 \\
      Always-plan + component & -3.20 & -1.78 & +0.40 & +0.45 & -2.67 & +1.47 & \textbf{-12.32} & -1.16 & \textbf{-0.35} & -1.69 & +0.11 \\
      Always-plan & -2.84 & -2.04 & -0.23 & \textbf{+0.10} & +0.45 & +1.44 & -12.23 & -1.52 & +0.50 & \textbf{-1.89} & \textbf{-0.12} \\
      \algo{} (Ours) & \textbf{-3.36} & \textbf{-3.19} & \textbf{-0.82} & +0.45 & \textbf{-3.80} & +1.40 & -11.60 & \textbf{-2.25} & +0.12 & -1.32 & -0.04 \\
      \bottomrule
    \end{tabular}}
\end{table*}

\begin{table*}[!htbp]
  \centering
  \caption{Synthetic ablation test gaps (\%).}
  \label{tab:ablation_full_test}
  \setlength{\tabcolsep}{3pt}
  \renewcommand{\arraystretch}{1.12}
  \resizebox{\textwidth}{!}{\begin{tabular}{l*{11}{r}}
      \toprule
      Method & C-CVRP & U-CVRP & TSP & BP & OP & Decap & JSS & PFSS & RCPS & Max-Cut & MIS \\
      \midrule
      Fixed plan & -3.26 & -1.81 & +2.53 & +0.69 & +7.97 & +2.66 & -6.30 & -0.50 & +0.39 & -0.86 & +1.20 \\
      Adaptive plan only & -3.00 & -1.81 & +2.44 & +0.69 & +3.76 & +2.01 & \textbf{-6.57} & -0.64 & +0.65 & -0.79 & +0.90 \\
      Component only & -3.47 & -1.59 & +4.17 & +0.69 & +3.86 & +1.62 & -6.41 & -0.69 & +1.61 & -0.70 & +1.79 \\
      Earlier full configuration & -3.45 & -1.06 & +1.78 & +0.69 & +7.04 & +1.98 & -6.42 & -0.89 & +0.57 & \textbf{-1.54} & +0.57 \\
      Always-plan + component & -3.62 & +0.09 & +1.97 & +0.69 & \textbf{-2.39} & \textbf{+1.82} & -6.19 & -0.93 & +0.43 & -1.08 & +1.52 \\
      Always-plan & -2.95 & -1.96 & +0.18 & \textbf{+0.38} & +8.12 & +2.03 & -6.17 & -1.67 & +0.43 & -1.44 & \textbf{+0.11} \\
      \algo{} (Ours) & \textbf{-4.38} & \textbf{-3.48} & \textbf{-0.70} & +0.69 & -1.94 & +1.90 & -5.52 & \textbf{-1.85} & \textbf{0.00} & -0.85 & +0.14 \\
      \bottomrule
    \end{tabular}}
\end{table*}

\subsection{Island initialization ablations}
\label{app:island_ablations}

Tables~\ref{tab:island_full_val} and~\ref{tab:island_full_test} report objective gaps of every problem for the \emph{Single island}, \emph{Shared plan}, and \emph{Partitioned library} variants in Table~\ref{tab:island_initialization}.  
Compared with \emph{Single island}, \emph{Shared plan} achieves lower test gaps on six tasks, including a reduction from $+2.09\%$ to $-2.96\%$ on U-CVRP, showing that multiple islands can help even when initialized with the same plan. With the island count fixed at five, \algo{} uses distinct initial plans and improves upon \emph{Shared plan} on seven validation tasks. It also lowers the test gaps on U-CVRP, OP, and RCPS, with the OP gap decreasing from $+3.85\%$ to $-1.94\%$. Although individual tasks favor different configurations, these results support the importance of both the number of islands and the diversity of their initial plans.

\begin{table*}[!htbp]
  \centering
  \caption{Island ablation validation gaps (\%). Bold indicates the lowest gap in each column, including ties.}
  \label{tab:island_full_val}
  \setlength{\tabcolsep}{3pt}
  \renewcommand{\arraystretch}{1.12}
  \resizebox{\textwidth}{!}{\begin{tabular}{l*{11}{r}}
      \toprule
      Variant & C-CVRP & U-CVRP & TSP & BP & OP & Decap & JSS & PFSS & RCPS & Max-Cut & MIS \\
      \midrule
      Single island & -2.88 & +3.10 & -0.92 & +0.49 & -0.06 & \textbf{+1.05} & -7.07 & \textbf{-2.65} & -0.57 & -1.32 & +0.10 \\
      Shared plan & -2.69 & -1.93 & \textbf{-0.98} & +0.62 & -0.97 & +1.55 & -6.61 & -2.35 & +1.21 & \textbf{-1.41} & \textbf{-0.10} \\
      Partitioned library & -1.88 & -1.94 & -0.93 & +0.49 & \textbf{-3.98} & +1.17 & -6.74 & -1.86 & \textbf{-1.78} & -1.23 & +0.10 \\
      \midrule
      \algo{} (Ours) & \textbf{-3.36} & \textbf{-3.19} & -0.82 & \textbf{+0.45} & -3.80 & +1.40 & \textbf{-11.60} & -2.25 & +0.12 & -1.32 & -0.04 \\
      \bottomrule
    \end{tabular}}
\end{table*}

\begin{table*}[!htbp]
  \centering
  \caption{Island ablation test gaps (\%). Bold indicates the lowest gap in each column, including ties.}
  \label{tab:island_full_test}
  \setlength{\tabcolsep}{3pt}
  \renewcommand{\arraystretch}{1.12}
  \resizebox{\textwidth}{!}{\begin{tabular}{l*{11}{r}}
      \toprule
      Variant & C-CVRP & U-CVRP & TSP & BP & OP & Decap & JSS & PFSS & RCPS & Max-Cut & MIS \\
      \midrule
      Single island & \textbf{-4.87} & +2.09 & -0.79 & \textbf{+0.50} & +4.18 & +1.20 & -6.90 & \textbf{-2.31} & +0.29 & \textbf{-1.60} & 0.00 \\
      Shared plan & -4.55 & -2.96 & \textbf{-0.84} & \textbf{+0.50} & +3.85 & \textbf{+1.06} & \textbf{-7.46} & -2.05 & +0.58 & -1.42 & \textbf{-0.26} \\
      Partitioned library & -3.58 & \textbf{-3.74} & -0.78 & \textbf{+0.50} & -1.46 & +1.45 & -6.90 & -1.52 & +0.29 & -1.51 & -0.07 \\
      \midrule
      \algo{} (Ours) & -4.38 & -3.48 & -0.70 & +0.69 & \textbf{-1.94} & +1.90 & -5.52 & -1.85 & \textbf{0.00} & -0.85 & +0.14 \\
      \bottomrule
    \end{tabular}}
\end{table*}

\newpage

\section{Dataset Details}
\label{app:datasets}

\subsection{Dataset Composition}

We consider three data types: Synthetic Problems, MIPLIB-Derived Problems,
and Nonlinear Geometry Benchmarks. The first two form the instance collections
summarized below; the nonlinear configurations are described in
Section~\ref{app:nonlinear_datasets}.
In these two collections, a task denotes one configured synthetic problem or
one MIPLIB-NL base problem together with its perturbations.
The 11 synthetic tasks represent 10 mathematical problem families because
capacitated vehicle routing has two configurations. The six MIPLIB-derived
tasks cover four application domains.

Table~\ref{tab:collections} counts the instances used in the reported experiments.
Each synthetic task uses 11 instances: \datasetid{seed0}--\datasetid{seed4}
for training, \datasetid{seed5}--\datasetid{seed6} for validation, and
\datasetid{seed7}--\datasetid{seed10} for testing. The MIPLIB experiments
reported here use five training instances per task, \datasetid{seed0}--\datasetid{seed4}.
These are instance seeds, rather than independent seeds of heuristic-evolution
runs. Section~\ref{app:split_scope} qualifies the effective diversity of the
decap-placement task.

\paragraph{Instance Format.}
All our data follow the MIPLIB-NL format \cite{li2026constructing}.
Each instance includes a natural-language problem description, large input
data supplied as CSV tables, and an initial Gurobi program encoding the
mathematical formulation.
A standardized solution-output format and a task-specific validator allow
generated programs to be evaluated consistently: the validator checks the
output solution's feasibility and the consistency of its reported objective.

\begin{table}[htbp]
  \centering
  \small
  \caption{Instances used in the reported experiments. Dashes indicate splits outside the reported evaluation.}
  \label{tab:collections}
  \begin{tabular}{@{}lrrrrr@{}}
    \toprule
    Collection & Tasks & Train & Val. & Test & Total \\
    \midrule
    Synthetic Problems & 11 & 55 & 22 & 44 & 121 \\
    MIPLIB-Derived Problems & 6 & 30 & -- & -- & 30 \\
    \midrule
    Total & 17 & 85 & 22 & 44 & 151 \\
    \bottomrule
  \end{tabular}
\end{table}

\paragraph{Data Availability.}
We will soon open-source both our synthetic dataset and our MIPLIB-NL-derived
dataset, together with the generation code, instance specifications, reference
models, and split definitions.

\paragraph{Fitness Score Function Definition.}
We use task-specific fitness functions to evaluate candidate programs. As defined earlier in Section~\ref{sec:hierarchy}, $\mathcal{D}$ denotes the training set for a target task, and $d \in \mathcal{D}$ represents an individual instance. Let $\operatorname{obj}(G;d)$ denote the verified objective value of the feasible solution returned by program $G$ on instance $d$ within the runtime budget $B$. For the synthetic and MIPLIB-derived problems (Tables~\ref{tab:synthetic_tasks} and~\ref{tab:miplib_tasks}), we define
\begin{equation}
    f(G;d) = -\frac{\left|\operatorname{obj}(G;d)-b_{48\mathrm{h}}(d)\right|}{\left|\operatorname{obj}(G;d)\right|+\epsilon},
\end{equation}
where $b_{48\mathrm{h}}(d)$ is the best objective bound obtained from a Gurobi reference run with a 48-hour time limit. This is a lower bound for minimization problems and an upper bound for maximization problems, rather than the objective value of Gurobi's best feasible solution. The bound is fixed for each instance across program evaluations, and $\epsilon>0$ prevents division by zero.

For the nonlinear geometry problems (Appendix~\ref{app:nonlinear_datasets}), we directly transform the objective according to its optimization direction, 
\begin{equation}
    f(G;d)=-s(d)\operatorname{obj}(G;d),
    \qquad
    s(d)=
    \begin{cases}
        +1, & \text{for minimization},\\
        -1, & \text{for maximization}.
    \end{cases}
\end{equation}
Evolution maximizes the mean training fitness, $F(G)=|\mathcal{D}|^{-1}\sum_{d\in\mathcal{D}}f(G;d)$. Each nonlinear configuration is optimized separately, so its training set contains a single instance. These fitness scores guide program evolution and are distinct from the signed gaps relative to Gurobi's 48-hour incumbent reported in the experimental results. 

\subsection{Synthetic Problems}
\label{app:synthetic_tasks}
\label{app:synthetic_generation}
\label{app:split_scope}

Table~\ref{tab:synthetic_tasks} describes the 11 synthetic tasks and how we generate their instances. We follow published sampling recipes where noted
and specify our changes in the corresponding entries. Problem
sizes are fixed within a task; realized edge counts, precedence counts, and
demand-derived fleet sizes can vary. Numerical ranges below are measured
across all 11 stored instances per task. Integer-uniform ranges include both
endpoints. For TSP, both CVRP configurations, and orienteering, coordinates
are stored to six decimal places, and distances between stored points $p_i$ and
$p_j$ are
\begin{equation}
  d_{ij}=\operatorname{round}\!\left(1000\lVert p_i-p_j\rVert_2\right).
  \label{eq:routing_distance}
\end{equation}

Decap placement requires a further qualification. Its randomized target
weights cancel against inversely scaled base requirements, leaving essentially
the same optimization model across seeds, apart from small rounding differences.
Its nominal split therefore does not measure generalization to substantively
different instances.

\begin{longtable}{@{}L{0.19\textwidth}L{0.77\textwidth}@{}}
  \caption{Synthetic problem definitions and generation settings.}
  \label{tab:synthetic_tasks}\\
  \toprule
  Task & Definition, generation details, and references \\
  \midrule
  \endfirsthead
  \multicolumn{2}{@{}l}{\tablename~\thetable\ (continued)}\\
  \toprule
  Task & Definition, generation details, and references \\
  \midrule
  \endhead
  \midrule
  \multicolumn{2}{r@{}}{Continued on next page}\\
  \endfoot
  \bottomrule
  \endlastfoot

  \taskname{Clustered CVRP} &
  Minimize total distance while serving
  each customer exactly once using exactly 22 depot-return routes, each carrying
  at most 55 demand units.
  \par\smallskip
  There are 120 customers in four Gaussian clusters
  with coordinate standard deviation 0.08, clipped to the unit square. The depot
  is at $(0.1,0.1)$; customer demands are uniform integers from 1 to 13.
  Each customer independently chooses a cluster center uniformly from
  $(0.25,0.25)$, $(0.75,0.30)$, $(0.35,0.78)$, and $(0.78,0.75)$ before its
  coordinates are sampled.
  \par\smallskip
  This is our custom clustered generator for the
  CVRP of \citet{dantzig1959truck}, not a reproduction of a published instance set. \\
  \addlinespace

  \taskname{Uniform CVRP} &
  Minimize total distance with
  complete single-visit service, depot-return routes, and vehicle capacity 50.
  \par\smallskip
  The 200 customers are sampled uniformly from the unit square, with a depot
  at $(0.5,0.5)$ and demands uniform on the integers 1--9. The fleet size is
  fixed for each instance and ranges from 20 to 22 vehicles:
  \[
    K=\max\!\left(\left\lceil\frac{\sum_i q_i}{50}\right\rceil+1,
    K_{\mathrm{FFD}}\right),
  \]
  where $K_{\mathrm{FFD}}$ is the number of capacity-50 bins obtained by
  first-fit decreasing on the demands.
  \par\smallskip
  Coordinates, demands, depot, and
  capacity follow the ACO benchmark sampling of ReEvo \cite{ye2024reevo};
  the exact fleet-size constraint and integer distances are our adaptations. \\
  \addlinespace

  \taskname{Decap-Placement Proxy} &
  Minimize capacitor cost subject to placing at
  most one capacitor type at each site and meeting every target's required
  cumulative synthetic effect.
  \par\smallskip
  A $30\times30$ grid supplies 900 sites and
  900 targets. Four types give 3,600 site--type choices; effects decay with
  distance inside a type-specific radius. Type $k$ contributes
  $e_k/(1+d_{st})$ when the Euclidean grid distance $d_{st}\leq\rho_k$, and zero otherwise.
  The four $(\mathrm{cost},e_k,\rho_k)$ tuples are $(1.0,1.4,1.25)$,
  $(2.1,2.6,2.25)$, $(3.8,4.3,3.5)$, and $(6.4,7.0,5.25)$.
  \par\smallskip
  Before rounding, target $t$ requires $0.22m_tF_t$, where $F_t$ is the
  strongest-type all-site coverage and
  $m_t=0.90+0.20\lVert p_t-(14.5,14.5)\rVert_2/(14.5\sqrt{2})$.
  Target-weight jitter is uniform on $[-0.06,0.06]$ but cancels against
  inversely scaled base requirements, up to rounding.
  This covering proxy is inspired by ReEvo's application \cite{ye2024reevo},
  not its electrical simulator or data distribution; see
  Section~\ref{app:split_scope} for the diversity qualification. \\
  \addlinespace

  \taskname{Job-Shop Scheduling} &
  \datasetid{job_shop_scheduling}. Minimize makespan while preserving each
  job's operation order and preventing machine conflicts. Operations cannot
  be interrupted.
  \par\smallskip
  There are 100 jobs and 20 machines (2,000 operations).
  Every job visits every machine once in a random order, with processing times
  uniform on the integers 1--99. This follows Taillard's random-permutation
  and processing-time recipe \cite{taillard1993benchmarks}, with fresh seeds;
  the scheduling formulation follows \citet{manne1960jobshop}. \\
  \addlinespace

  \taskname{Max-Cut} &
  \datasetid{max_cut}. Partition the vertices into two sets to maximize the
  number of crossing edges \cite{goemans1995improved}.
  \par\smallskip
  Each graph has 2,000 vertices and
  10,882--11,072 edges. Our hybrid generator combines Barab\'asi--Albert-style
  preferential attachment \cite{barabasi1999emergence}, initialized with a
  five-vertex clique and attachment parameter 4, with an independent
  Erd\H{o}s--R\'enyi $G(n,p)$ edge overlay at $p=0.0015$
  \cite{erdos1959random,gilbert1959random}. Duplicate edges are merged;
  all stored edge weights equal one. This is our custom combination of the
  two graph models. \\
  \addlinespace

  \taskname{Maximum Independent Set} &
  \datasetid{maximum_independent_set}. Maximize the cardinality of a vertex
  subset containing no adjacent pair.
  \par\smallskip
  Each preferential-attachment graph has
  1,500 vertices and 11,964 edges, uses attachment parameter 8, and starts from
  a nine-vertex clique. We follow the preferential-attachment principle of
  \citet{barabasi1999emergence} with this explicit initialization and use
  the unweighted independent-set objective \cite{nemhauser1975vertex}. \\
  \addlinespace

  \taskname{Offline Bin Packing (BP)} &
  \datasetid{offline_bin_packing}. Minimize the number of bins while assigning
  each item to exactly one bin without exceeding capacity. All items are
  available before packing \cite{martello1990lower}.
  \par\smallskip
  Each instance has 1,000 items with independently
  sampled uniform integer sizes from 20 to 100 and bin capacity 150, following
  Falkenauer's uniform benchmark recipe \cite{falkenauer1996hybrid}.
  These are newly generated instances, not the original benchmark files. \\
  \addlinespace

  \taskname{Orienteering} &
  \datasetid{orienteering}. Maximize the prize collected on one depot-return
  tour visiting a subset of customers within a travel-distance budget.
  \par\smallskip
  The 500 nodes comprise a centered depot and 499 uniformly distributed customers.
  The distance budget is $\operatorname{round}(1000\cdot0.4\sqrt{500})=8944$.
  Customer prizes are
  \[
    \pi_i=\operatorname{clip}_{[1,100]}\!\left(
      \operatorname{round}(15+75r_i+\epsilon_i)\right),
  \]
  where $\epsilon_i\sim\mathcal N(0,8^2)$ and $r_i$ is the distance from the
  centered depot divided by $\sqrt{0.5}$, computed before coordinate rounding.
  \par\smallskip
  The distance-correlated prize principle follows ReEvo \cite{ye2024reevo};
  noise, normalization, centered depot, and budget are our adaptations to the
  orienteering problem \cite{golden1987orienteering}. \\
  \addlinespace

  \taskname{Permutation Flow-Shop Scheduling} &
  \datasetid{permutation_flow_shop_scheduling}. Minimize makespan. All jobs
  follow the same machine order, and all machines process jobs in the same
  chosen permutation. Operations of a job cannot overlap, and each machine
  processes at most one job at a time.
  \par\smallskip
  The 100 jobs and 20 machines have
  2,000 independently sampled uniform integer processing times from 1 to 100.
  This is a Taillard-style synthetic recipe \cite{taillard1993benchmarks},
  using an upper endpoint of 100 rather than Taillard's 99; we do not reuse
  the published benchmark instances. \\
  \addlinespace

  \taskname{Resource-Constrained Project Scheduling} &
  \datasetid{resource_constrained_project_scheduling}. Minimize makespan subject
  to finish--start precedence and renewable-resource capacities during each
  activity's uninterrupted execution.
  \par\smallskip
  Instances have 85 activities, five
  resources, and 212--265 precedence arcs, with no dummy activities.
  Durations are uniform integers
  from 1 to 8 and capacities from 5 to 12. The first resource has comparatively
  larger capacity-relative demands: activity demands are uniform integers
  from zero to $\lfloor C_0/2\rfloor$ for resource zero and to
  $\lfloor C_r/3\rfloor$ otherwise.
  \par\smallskip
  Each pair $i<j$ receives an arc with
  probability 0.06; consecutive pairs receive an independent additional
  opportunity with probability 0.35, with duplicates merged. This is our
  custom acyclic-network recipe for the classical formulation
  \cite{pritsker1969multiproject}, not a replication of PSPLIB generation. \\
  \addlinespace

  \taskname{Traveling Salesman} &
  \datasetid{traveling_salesman}. Minimize the length of a Hamiltonian cycle
  visiting every node exactly once and returning to its start.
  \par\smallskip
  Each instance
  contains 350 nodes drawn from three Gaussian clusters with coordinate
  standard deviation 0.08 and clipped to the unit square. Each node independently
  chooses one of $(0.2,0.25)$, $(0.78,0.24)$, and $(0.5,0.75)$ with equal
  probability. This is our custom clustered Euclidean generator; the reference
  model uses the subtour formulation of \citet{miller1960integer}. \\
\end{longtable}

\subsection{MIPLIB-Derived Problems}
\label{app:miplib_tasks}

Table~\ref{tab:miplib_tasks} describes the six MIPLIB-derived tasks.
MIPLIB-NL \cite{li2026constructing} reconstructs natural-language optimization
problems from MIPLIB~2017 \cite{gleixner2021miplib}. We selected
these six base problems because Gurobi \cite{gurobi} did not certify optimality within
two hours (7,200 seconds) per instance. This selection criterion refers to
the initial screening, not subsequent reference re-solves or a claim about
all instances in MIPLIB-NL.
The tasks comprise two curriculum-based course-scheduling problems
(\datasetid{comp12-2idx}, \datasetid{comp21-2idx}) \cite{lach2012curriculum},
two graph-drawing problems (\datasetid{graphdraw-grafo2},
\datasetid{graphdraw-mainerd}) \cite{e2017drawing}, mining-project selection
(\datasetid{opm2-z12-s8}), and PCB assembly-line configuration
(\datasetid{sct32}) from MIPLIB \cite{koch2011miplib}.
We use these original IDs consistently in the dataset and result tables.
For each task, the five training instances comprise the unperturbed reconstructed
base instance (\datasetid{seed0}) and four task-specific perturbations
(\datasetid{seed1}--\datasetid{seed4}). The dimensions below are fixed across
these training instances.

\begin{longtable}{@{}L{0.27\textwidth}L{0.69\textwidth}@{}}
  \caption{MIPLIB-derived problem definitions.}
  \label{tab:miplib_tasks}\\
  \toprule
  Problem ID & Definition and instance configuration \\
  \midrule
  \endfirsthead
  \multicolumn{2}{@{}l}{\tablename~\thetable\ (continued)}\\
  \toprule
  Problem ID & Definition and instance configuration \\
  \midrule
  \endhead
  \midrule
  \multicolumn{2}{r@{}}{Continued on next page}\\
  \endfoot
  \bottomrule
  \endlastfoot

  \datasetid{comp12-2idx} &
  \datasetid{Course scheduling.} Schedule all required lectures into available periods
  while avoiding course conflicts and respecting the number of rooms. Minimize
  penalties for room-capacity shortfalls, insufficient teaching-day spread,
  and isolated curriculum lectures.
  \par\smallskip
  There are 88 courses, 218 lectures,
  six days with six periods each, 11 rooms, 146 curricula, and 519
  course-conflict pairs. \\
  \addlinespace

  \datasetid{comp21-2idx} &
  \datasetid{Course scheduling.} Solve the same time-assignment problem and three-part
  penalty objective, with a different timetable, room inventory, and curriculum
  structure. Room capacities are represented by capacity thresholds and
  penalized shortfalls.
  \par\smallskip
  There are 94 courses, 327 lectures, five days with
  five periods each, 18 rooms, 78 curricula, and 302 conflict pairs. \\
  \addlinespace

  \datasetid{graphdraw-grafo2} &
  \datasetid{Graph drawing.} Place entities at integer coordinates inside
  their permitted windows, enforcing size-dependent non-overlap and consistent
  directional relations. Minimize weighted connection-distance terms plus
  deviation from preferred anchors.
  \par\smallskip
  This is the entity-placement phase of
  graph drawing, with 67 entities, 73 connections, and 2,211 entity pairs.
  All edge weights equal 122. \\
  \addlinespace

  \datasetid{graphdraw-mainerd} &
  \datasetid{Graph drawing.} Solve the same entity-placement problem on a
  smaller graph, with placement windows, pairwise separations, directional
  consistency, and centering terms.
  \par\smallskip
  There are 31 entities, 33 connections,
  and 465 entity pairs. All edge weights equal 126. \\
  \addlinespace

  \datasetid{opm2-z12-s8} &
  \datasetid{Mining-project selection.} Select mining blocks/projects to maximize total net
  present value. Selecting a project requires selecting its predecessors,
  and the total use of each resource must respect its capacity.
  \par\smallskip
  There are 10,800 binary project decisions, eight resources, and 319,500 dependency
  arcs. Every resource has capacity 3,704,832. \\
  \addlinespace

  \datasetid{sct32} &
  \datasetid{PCB assembly-line configuration.} Assign each operation's full workload across eligible
  workstation options, using binary decisions on discrete lanes and fractional
  assignments where permitted. Minimize staffing cost minus weighted checkpoint
  performance and discrete-routing rewards, subject to capacities, checkpoint
  constraints, priority-token eligibility and budget, and policy checks.
  \par\smallskip
  There are 552 operations, ten workstation options, seven board variants,
  27 checkpoints, 67 policy checks, and a priority-token budget of six. \\
\end{longtable}

\subsubsection{Generating MIPLIB Training Instances}
\label{app:miplib_perturbations}

We construct the five training instances for each of the six reconstructed
base problems as follows. For each problem, \datasetid{seed0} retains the base data,
and \datasetid{seed1}--\datasetid{seed4} vary selected relationships and
coefficients while preserving table dimensions, course--capacity keys, or
graph degree sequences, as appropriate. For \datasetid{sct32}, we use the
upstream generator's numerical-jitter mode. Across all six tasks, the
variable count remains unchanged; linear constraint counts may vary for
timetabling (Table~\ref{tab:all_formulation_sizes}).

\paragraph{Course Timetabling.}
For both tasks, we relocate unavailable periods while preserving their total
count for each course. We first sample full-day unavailability and then
remaining slots without replacement, using the base instance's corresponding
day and day--period frequencies plus one as sampling weights. We retain
the existing curriculum-induced conflict pairs and apply
degree-preserving double-edge swaps to the remaining conflicts: 83 pairs
for \datasetid{comp12-2idx} and 32 for \datasetid{comp21-2idx}. We also permute
lecture-count/minimum-day pairs across courses, preserving their multiset
and total lecture count. A permutation is accepted only if each course's
lecture count fits its available periods and each curriculum's lecture load
does not exceed the base instance's maximum curriculum load. Finally, we
permute capacity-penalty weights within each capacity class. Curricula,
room inventories, course identifiers, and course--capacity penalty keys
are retained.

\paragraph{Graph Drawing.}
Degree-preserving edge swaps modify the connection graph while retaining
entity data, placement windows, sizes, edge count, degree sequence, and the
constant edge weight. Edge-distance coefficients are then recomputed from the
unchanged geometry.

\paragraph{Mining-Project Selection.}
Dependency arcs and resource capacities are retained. Project values and
resource costs are independently multiplied by factors from $U(0.9,1.1)$
and rounded to integers; stored resource-cost entries are bounded below by
one, while omitted entries remain zero. Thus,
dependency structure is preserved, while changes to resource coefficients
can change the feasible set.

\paragraph{PCB Assembly-Line Configuration.}
The upstream generator's jitter mode independently multiplies entries in
designated numeric data-table columns by factors from $U(0.9,1.1)$, using
floating-point conversion and nonnegative clipping without integer rounding.
These columns contain capacities, workloads, performance coefficients,
weights, and constraint right-hand sides. Identifiers, categorical flags,
table structure, and scalar model parameters are retained.

\paragraph{Feasibility Checks.}
We check the timetable variable counts against an explicit counting formula
and verify that the graph-drawing coefficient formulas reproduce the base
data. We then construct each instance's Gurobi model to record its size before
optimization, obtain a reference solution, and pass the reported solution to
the task-specific verifier to check feasibility and objective consistency.
Generation-time reference runs used a 120-second budget with two threads,
with additional 600-second, four-thread runs for \datasetid{graphdraw-grafo2}.
The generation-time verification records contain a \datasetid{VALID} verdict
for each of the 30 training instances; these checks are separate from both
the initial hardness screening and subsequent longer reference re-solves.
The validated reference objectives are feasible-solution values; a successful
verification does not certify optimality. The timetabling guards are screening
conditions, not a guarantee of feasibility. If a draw is proven infeasible, we resample it
using a new random-stream attempt index and repeat validation. 

Seeds, resampling attempts, and perturbation details are recorded with each
instance, together with its feasibility-check results.

\subsubsection{MIPLIB Model Sizes}
\label{app:formulation_sizes}

Table~\ref{tab:formulation_sizes} reports the numbers of decision variables and constraints in the Gurobi models used for the six MIPLIB-derived problems. Each problem
has the same number of variables across its 11 instances. General constraints
are reported separately from linear constraints.

\begingroup
\small
\setlength{\tabcolsep}{3pt}
\begin{longtable}{@{}L{0.25\textwidth}rrrL{0.29\textwidth}@{}}
\caption{Formulation sizes before presolve for 11 synthetic and six MIPLIB-derived tasks. Ranges cover synthetic instances; ``varies'' marks MIPLIB constraint counts affected by perturbation.}
\label{tab:all_formulation_sizes}
\label{tab:formulation_sizes}
\\\toprule
Problem & Variables & \shortstack{Linear\\constraints} & \shortstack{General\\constraints} & Key instance parameters \\
\midrule
\endfirsthead
\multicolumn{5}{@{}l}{\tablename~\thetable\ (continued)}\\
\toprule
Problem & Variables & \shortstack{Linear\\constraints} & \shortstack{General\\constraints} & Key instance parameters \\
\midrule
\endhead
\midrule
\multicolumn{5}{r@{}}{Continued on next page}\\
\endfoot
\bottomrule
\endlastfoot
C-CVRP & 14,640 & 14,522 & 0 & 120 customers; 22 vehicles; capacity 55 \\
U-CVRP & 40,400 & 40,202 & 0 & 200 customers; 20--22 vehicles; capacity 50 \\
OP & 250,498 & 250,501 & 0 & 500 nodes; travel budget 8,944 \\
TSP & 122,499 & 122,152 & 0 & 350 nodes \\
JSSP & 101,001 & 201,900 & 0 & 100 jobs; 20 machines; 2,000 operations \\
PFSSP & 12,000 & 4,081 & 0 & 100 jobs; 20 machines \\
RCPSP & 28,646--34,358 & 2,140--2,449 & 0 & 85 activities; 5 resources; 212--265 precedences \\
BP & 1,001,000 & 2,000 & 0 & 1,000 items; bin capacity 150 \\
\datasetid{comp12-2idx} & 11,726 & varies & 0 & 88 courses; 36 time slots; 11 rooms; 519 conflicts \\
\datasetid{comp21-2idx} & 10,911 & varies & 0 & 94 courses; 25 time slots; 18 rooms; 302 conflicts \\
Max-Cut & 12,882--13,072 & 43,528--44,288 & 0 & 2,000 vertices; 10,882--11,072 edges \\
MIS & 1,500 & 11,964 & 0 & 1,500 vertices; 11,964 edges \\
\datasetid{graphdraw-grafo2} & 9,258 & 194,611 & 9,136 & 67 entities; 73 relationships; 2,211 entity pairs \\
\datasetid{graphdraw-mainerd} & 2,050 & 18,801 & 1,992 & 31 entities; 33 relationships; 465 entity pairs \\
DCP & 3,600 & 1,800 & 0 & 900 sites; 900 targets; 4 capacitor types \\
\datasetid{opm2-z12-s8} & 10,800 & 319,508 & 0 & 10,800 projects; 8 resources; 319,500 dependencies \\
\datasetid{sct32} & 6,607 & 3,885 & 0 & 552 operations; 10 stations; 7 variants; 27 checkpoints \\
\end{longtable}
\endgroup

\subsection{Nonlinear Geometry Benchmarks}
\label{app:nonlinear_datasets}

We take our nonlinear geometry benchmarks from \citet{berthold2026global},
who revisit problems studied in AlphaEvolve
\cite{georgiev2025mathematical,novikov2025alphaevolve} by solving those with nonlinear programming solvers. We retain these
AlphaEvolve problem definitions and use the nonlinear formulations presented
by \citet{berthold2026global}. We manually wrote our own natural-language
problem descriptions and mathematical formulation files for these settings.
We evolve heuristics separately for each problem--size configuration.

Circle packing maximizes the sum of variable radii in either a unit square
or a rectangle of perimeter four with an optimized aspect ratio. The
distance-ratio problem minimizes the ratio of maximum to minimum pairwise
distances between points; our formulation uses squared distances and fixes
the minimum squared distance to one. Hexagon packing minimizes the side
length of a regular hexagonal container holding non-overlapping unit-side
regular hexagons with free translations and rotations.

Table~\ref{tab:nonlinear_configurations} lists all 14 configurations in the nonlinear collection. Here, $n$ denotes the number of circles, points, or hexagons, and $d$ denotes the dimension of the point configuration.

\begin{table}[htbp]
  \centering
  \small
  \renewcommand{\arraystretch}{1.2}
  \caption{Nonlinear problem names and size configurations.}
  \label{tab:nonlinear_configurations}
  \begin{tabular}{@{}L{0.48\textwidth}L{0.48\textwidth}@{}}
    \toprule
    Problem Name & Size Configurations \\
    \midrule
    \datasetid{circle_packing_square} & $n\in\{26,32\}$ \\
    \datasetid{circle_packing_rectangle} & $n\in\{21,26,27\}$ \\
    \datasetid{distance_ratio} & $d=2,\ n\in\{16,21,22\}$;\newline
      $d=3,\ n=14$ \\
    \datasetid{hexagon_packing} & $n\in\{11,12,14,15,16\}$ \\
    \bottomrule
  \end{tabular}
\end{table}

These fixed geometry configurations do not use separate instances for
training, validation, and testing. Seed directories identify problem sizes
rather than random samples. These benchmarks are additional to the 151
instances used in the reported experiments in Table~\ref{tab:collections}.

\section{Additional Method Details}
\label{app:method-details}

\setcounter{equation}{0}
\renewcommand{\theequation}{\Alph{section}\arabic{equation}}
\setcounter{table}{0}
\renewcommand{\thetable}{\Alph{section}\arabic{table}}
\setcounter{algorithm}{0}
\renewcommand{\thealgorithm}{\Alph{section}\arabic{algorithm}}
\ifdefined\theHequation
  \renewcommand{\theHequation}{supp.\Alph{section}.\arabic{equation}}
\fi
\ifdefined\theHtable
  \renewcommand{\theHtable}{supp.\Alph{section}.\arabic{table}}
\fi
\ifdefined\theHalgorithm
  \renewcommand{\theHalgorithm}{supp.\Alph{section}.\arabic{algorithm}}
\fi

\subsection{Notations}
\label{app:annotations}

The indices $k,j,h,i$ identify the island, global round, local update, and program step, respectively. The tag $\mathrm{role}$ is $\mathrm{seed}$, $\mathrm{par}$, $\mathrm{child}$, or $\star$. Omit $h$ when no local update is involved. For shared best programs, $\mathrm{scope}$ is $\mathrm{global}$ or $\mathrm{archive}$.

\begin{table}[!htbp]
\centering
\setlength{\belowcaptionskip}{5pt}
\caption{Annotations}
\label{tab:annotations}
\begingroup
\setlength{\tabcolsep}{5pt}
\renewcommand{\arraystretch}{1.25}
\begin{tabularx}{\linewidth}{@{}>{\raggedright\arraybackslash}p{0.2\linewidth}>{\raggedright\arraybackslash}X@{}}
\toprule
\textbf{Annotation} & \textbf{Description} \\
\midrule
$\mathcal A,\mathcal A_j$ & Shared Program Database of retained, validated programs. \\
$\mathcal A_j^{(k)}$ & All retained programs from island $k$, including earlier plans. \\
$\mathcal A_j^{(k,\mathrm{par})},\mathcal C_j^{(k)}$ & Parent-eligible programs under the current plan generation. \\
$\mathcal A_j^{(k,\mathrm{ins})}$ & Available inspiration programs for island $k$. \\
$\mathcal D,B$ & Training set and per-instance execution budget. \\
$f(G;d),F(G)$ & Instance fitness and average training fitness; larger is better. \\
$F_{j,h}^{(k,\mathrm{role})}$ & Fitness of the correspondingly indexed program $G$; shared scopes use the same convention. \\
$g_i,G$ & Step implementation and complete program $G=g_1\odot\cdots\odot g_m$. \\
$\mathcal G(P)$ & Programs preserving the ordered subgoals and components of $P$. \\
$G_{j,h}^{(k,\mathrm{role})}$ & Seed, parent, child, or current-plan incumbent, as indicated by the role tag. \\
$G_j^{(\mathrm{scope},\star)}$ & Best current-island incumbent (global) or best retained program (archive); add $h$ for local updates. \\
$H_{j,h}^{(k)}$ & Recent-progress estimate controlling plan review. \\
$\mathcal I_{j,h}^{(k)}$ & Inspiration programs actually sampled for an update. \\
$K,J,W$ & Island count, global-round count, and local code visits after an accepted plan proposal. \\
$M_{\mathrm{dbg}}$ & Debugging-attempt limit for bounded generation calls. \\
$n_j^{(k)},N_j$ & Island and total code-improvement visit counts; retries do not create extra visits. \\
$P_j^{(k)}$ & Plan associated with island $k$. \\
$Q_j(s)$ & Mean component contribution since its last description change. \\
$R_{j,h}^{(k)},r_{j,h}^{(k)}$ & Discounted accumulated reward and current-visit reward. \\
$\mathcal R_j$ & Existing components selected for review. \\
$s,\mathcal S_j,s_{j,i}^{(k)}$ & Component, shared library, and component selected at step $i$. \\
$u_{j,i}^{(k)},a_{j,i}^{(k)}$ & Step subgoal and preliminary implementation annotation. \\
$V_{j,h}^{(k)}$ & Discounted exposure statistic. \\
$\beta_{j,i}^{(k)},\zeta_{j,i}^{(k)}$ & Step time allocation and step type. \\
$\delta_{j,h}^{(k)}$ & Nonnegative relative improvement in a code visit. \\
$\Delta_{j,h,i}^{(k)},\boldsymbol{\Delta}_j(s)$ & Step contribution and component history since its last description change. \\
$\Phi_{j,h}^{(k)}$ & Interpreter report for a parent--child comparison. \\
$\bot$ & No valid program obtained; not a numerical fitness. \\
\bottomrule
\end{tabularx}
\endgroup
\end{table}

\subsection{Agent Description} \label{app:agents}

\paragraph{Component Proposer.}
The Component Proposer receives the target problem description and proposes 10--20 relevant high-level components. The descriptions should cover distinct approaches without duplicating components that differ only in implementation details. Each component should be specific enough to guide planning while supporting multiple concrete algorithms.

\paragraph{Planner.}
The Planner selects components from the library and organizes them into $K$ different plans, with one plan assigned to each island. Each plan contains an ordered sequence of steps. The \emph{subgoal and selected component} are fixed during ordinary code-level evolution. The \emph{implementation annotation, step type, and time allocation} are revisable implementation fields. The annotation describes a preliminary procedure that the Coder can follow directly. This preserves the distinction between the plan's structural specification and its implementation guidance. The Planner distinguishes three step types, listed below. It assigns time allocations according to these execution requirements and the shared per-instance budget.

\begin{itemize}[topsep=0pt,itemsep=0pt,parsep=0pt,partopsep=0pt,leftmargin=*]
    \item \textbf{Complete-algorithm.} Runs a prescribed finite procedure to completion according to its algorithmic stopping rule. Its runtime is estimated or measured based on the procedure and problem instance.
    \item \textbf{Budget-controlled.} Performs search within an allocated time limit. Its execution is controlled by a local runtime budget.
    \item \textbf{Hybrid.} Combines a required complete-algorithm component with budget-controlled search. It requires a minimum time allocation for the mandatory component, with additional time available for search.
\end{itemize}

\paragraph{Coder.}
The Coder translates a supplied plan into an initial executable program. The initial-generation prompt requires all plan steps to be implemented in order, following their annotations, step types, and runtime fields. Intermediate outputs are passed between steps. After each step, the program reports a solution and cumulative execution time, along with the step's objective value or gap and execution time. The Code Improver handles subsequent implementation changes, runtime reallocation, and permitted step deactivation.

\paragraph{Debugger.}
The Debugger handles programs rejected because of compilation errors, invalid solutions, time-limit exceedances, or other execution failures. It supports two modes. In \emph{direct repair}, the agent receives the rejected source code and its error message. It revises that implementation to address the reported failure while preserving the applicable plan requirements. In \emph{error-aware regeneration}, the agent generates replacement code from the original generation context, augmented with the error message. Rather than requiring a modification of the rejected implementation, this mode warns the agent against repeating the same failure during a new generation attempt. Every repaired or regenerated candidate is evaluated again. During evolution, debugging is bounded by $M_{\mathrm{dbg}}$. During initialization, seed generation continues until validation succeeds, and the main evolution procedure begins only after every island has a valid seed.

\paragraph{Code Improver.}
The Code Improver receives a parent program, inspiration programs, the current plan, the component library, and any available code-level revision suggestions. It generates a child while preserving the ordered subgoals and selected components. Its revisions may change concrete algorithms, data structures, parameters, implementation annotations, step types, stopping rules, time allocations, and interactions between adjacent steps. The parent and inspirations are selected by the system's \emph{SelectCode} function. 
The Code Improver may also deactivate a step judged unnecessary, harmful, or too expensive by assigning it zero runtime while retaining its step identifier, component assignment, subgoal, and step type. The reason for deactivation is recorded in the step metadata. When a component has been removed from the active library, the Code Improver also receives guidance to bypass the corresponding step. The Plan Improver remains responsible for deleting the recorded plan entry or replacing its assigned component. 

\paragraph{Plan Improver.}
The Plan Improver receives the selected island's current plan and programs, details of the best-performing plan, and sampled plans from other islands. These records include components, subgoals, implementation annotations, time allocations, and measured performance. The agent also receives the complete active component library and accumulated plan-level feedback. The proposed plan may add, remove, replace, or reorder steps and change their subgoals or selected components. It includes preliminary implementation procedures and time allocations for the Coder. If the current plan references a removed component, the proposal must delete the affected step or replace its component with an available alternative. Every proposal that produces a valid seed starts a new plan generation, even when its content matches the preceding plan.

\paragraph{Component Improver.}
The Component Improver reviews selected components using their contribution histories and aggregated Interpreter feedback. It may refine descriptions, remove ineffective components, or consider missing high-level components suggested by the Interpreter. Only existing components used by the selected island's current plan are eligible for revision or removal during that round. Existing components absent from the plan remain unchanged. Suggestions for genuinely new components are considered separately. A description revision clears that component's numerical contribution history. A removal deletes its description from the active library but does not immediately rewrite existing programs or plans. Subsequent code improvement
receives a suggestion to bypass the affected step, while subsequent plan improvement must delete or replace it.

\paragraph{Interpreter.} \label{app:interpreter}
The Interpreter compares the parent and child using their source code and implementation feedback. This feedback includes each step's objective value or gap, execution time, and relevant execution diagnostics. The actual optimization solutions, such as complete schedules or decision-variable assignments, are not included in the comparison context. The Interpreter explains measured outcomes rather than replacing the evaluator's assessment. Its report contains four components, detailed below. Code-level insights are associated with the parent and child. Plan-level reports accumulate for the current plan generation; reports from retired generations remain historical records rather than entering the new generation's active feedback pool. Component-level reports accumulate across relevant evaluations and islands. A proposed addition must describe a missing high-level search family rather than only an algorithm, neighborhood, parameter, or solver setting.
\begin{itemize}[topsep=0pt,itemsep=0pt,parsep=0pt,partopsep=0pt,leftmargin=*]
    \item \textbf{Change Summary:} Records the measured outcome as \emph{better}, \emph{worse}, or \emph{equivalent}. Each material change identifies the affected step, change type, parent and child behaviors, observed effect, and attribution confidence. The summary becomes part of the child's evolution history.

    \item \textbf{Code-level Insight:} Assesses implementation changes under the unchanged subgoals and components. The newly generated code is identified as \emph{effective}, \emph{regression}, \emph{neutral}, or \emph{not established}. To support the argument, the interpreter includes supporting evidence and a concrete implementation direction. These insights are available to later Code Improver calls.

    \item \textbf{Plan-level Insight:} Assesses step usefulness, component suitability, organization, and runtime allocation. Step assessments use \emph{helpful}, \emph{supporting}, \emph{implementation limited}, \emph{component mismatch}, \emph{subgoal unhelpful}, or \emph{not established}. Structural recommendations use \emph{revise}, \emph{add}, or \emph{remove}.

    \item \textbf{Component-level Insight:} Produces one record per unique component used by the plan, consolidating evidence when several steps share that component. It records observed implementations, transferable lessons, and proposed operations such as \emph{keep}, \emph{propose revise}, or
    \emph{propose add}.
\end{itemize}

\subsection{Initialization} \label{app:initialization}
Initialization constructs the component library, generates $K$ plans, and obtains one validated seed for every island. It then performs $W$ local code-improvement visits per island before global scheduling begins. Component descriptions remain unchanged during these initial local visits. Algorithm~\ref{alg:initialization} gives the initialization procedure.

\begin{algorithm}[t]
\caption{The Initialization Procedure of HeurEvo.}
\label{alg:initialization}
\small
\begin{algorithmic}[1]
\Require Problem description, $K$ islands, training set $\mathcal D$,
per-instance budget $B$, local steps $W\geq1$, and debugging limit
$M_{\mathrm{dbg}}$.
\State $\mathcal S_0\leftarrow\mathsf{ProposeComponents}(\text{problem}),
\quad\mathcal A\leftarrow\varnothing$
\State $\{P_0^{(k)}\}_{k=1}^{K}\leftarrow
\mathsf{CreatePlans}(\mathcal S_0,K,B)$
\State Initialize an empty contribution array for each $s\in\mathcal S_0$
\Statex \hspace{\algorithmicindent}\textit{// Seed generation}
\For{$k=1,\ldots,K$}
    \State $(G_0^{(k,\mathrm{seed})},F_0^{(k,\mathrm{seed})})
    \leftarrow\mathsf{CodePlan}(P_0^{(k)},\mathcal S_0,B)$
    \State Insert the validated seed into $\mathcal A$ and associate it with island $k$
    \State $\mathcal A_{0,0}^{(k,\mathrm{par})}
    \leftarrow\{G_0^{(k,\mathrm{seed})}\},\quad
    F_{0,0}^{(k,\star)}\leftarrow F_0^{(k,\mathrm{seed})}$
    \State $(R_{0,0}^{(k)},V_{0,0}^{(k)},H_{0,0}^{(k)},n_{0,0}^{(k)})
    \leftarrow(0,1,0,0)$
\EndFor
\Statex \hspace{\algorithmicindent}\textit{// Local code evolution}
\For{$k=1,\ldots,K$}
    \For{$h=1,\ldots,W$}
        \State $(G_{0,h}^{(k,\mathrm{par})},\mathcal I_{0,h}^{(k)})
        \leftarrow\mathsf{SelectCode}(\mathcal A_{0,h-1}^{(k,\mathrm{par})},\mathcal A)$
        \State $(G_{0,h}^{(k,\mathrm{child})},F_{0,h}^{(k,\mathrm{child})})
        \leftarrow\mathsf{ImproveCode}(k,G_{0,h}^{(k,\mathrm{par})},\mathcal I_{0,h}^{(k)})$
        \If{$G_{0,h}^{(k,\mathrm{child})}\neq\bot$}
            \State $\Phi_{0,h}^{(k)}\leftarrow
            \mathsf{Interpret}(G_{0,h}^{(k,\mathrm{par})},G_{0,h}^{(k,\mathrm{child})})$
            \State $\mathsf{UpdateState}(k,G_{0,h}^{(k,\mathrm{child})},
            F_{0,h}^{(k,\mathrm{child})},\Phi_{0,h}^{(k)})$
            \Comment{Eqs.~\ref{eq:island-reward}--\ref{eq:island-updates}}
        \Else
            \State Apply Eq.~\ref{eq:island-updates} with
            $r_{0,h}^{(k)}=\delta_{0,h}^{(k)}=0$; add no program
            \State Retain the parent-eligible collection and incumbent for the next local update
        \EndIf
    \EndFor
    \State Record the final plan, programs, feedback, and statistics of island $k$ at index $0$
\EndFor
\State $\mathcal A_0\leftarrow\mathcal A,\quad
N_0\leftarrow\sum_{k=1}^{K}n_0^{(k)}=KW$
\State \Return $K$ initialized islands, $\mathcal S_0$, and $\mathcal A_0$
\end{algorithmic}
\end{algorithm}

In Algorithm~\ref{alg:initialization}, \emph{CodePlan} and \emph{ImproveCode} include generation, evaluation, and debugging internally. Each successful call returns the validated program together with its average training fitness. Evaluation records remain available for subsequent interpretation.

During initialization, \emph{CodePlan} continues repairing or regenerating a seed until validation succeeds. \emph{ImproveCode} instead uses bounded debugging. If no valid child is obtained within that limit, it returns $(\bot,\bot)$. The rejected candidate is not inserted into the Program Database, and the local visit receives zero reward and zero relative improvement.

Each valid seed initializes its island's parent-eligible collection and incumbent. Before local code improvement, $R, V, H, n$ are initialized to $0, 1, 0, 0$, respectively. Seed generation does not count as code improvement.

Every scheduled \emph{ImproveCode} call counts as one visit, whether it succeeds immediately, requires debugging, or remains invalid. The successful update or the failure update increments $n$ exactly once. Consequently,
\begin{align*}
n_0^{(k)}=W,\qquad N_0=KW.
\end{align*}
The requirement $W\geq1$ ensures that every island has a positive visit count before the first UCB selection. Initialization is separate from the $J$ global scheduling rounds.

\subsection{Evolution}
\label{app:evolution}

\paragraph{Location of the deferred details.}
The exact scheduling rules and complete evolution algorithm deferred from Section~\ref{sec:evolution} are collected here. References below to the ``main-text'' selection or update rules refer to Eqs.~\ref{eq:ucb}--\ref{eq:island-updates} in this subsection. Initialization and its starting values remain in Appendix~\ref{app:initialization}; restart, timeout, and invalid-child handling are specified in the operator descriptions below.

\paragraph{UCB island-selection rule.}
At the beginning of global round $j$, the controller selects an island using its pre-round statistics:
\begin{equation}
\label{eq:ucb}
\begin{aligned}
k_j&=\arg\max_k\left[
\frac{R_{j-1}^{(k)}}{V_{j-1}^{(k)}}
+C\sqrt{\frac{\log N_{j-1}}{n_{j-1}^{(k)}}}
\right],\\
N_{j-1}&=\sum_{\ell=1}^{K}n_{j-1}^{(\ell)}.
\end{aligned}
\end{equation}
Here, $R$ and $V$ summarize discounted rewards and exposure. The count $n$ records scheduled code-improvement visits to one island, and $N$ totals these visits across islands. The coefficient $C$ controls exploration. The first term favors recent normalized progress; the second encourages exploration of under-visited islands. Counts include unsuccessful visits, not individual debugging attempts. Initialization with $W\geq1$ ensures positive counts before the first selection.

\paragraph{Normalized reward and relative improvement.}
Let $h$ index local code-improvement visits in round $j$, with $k=k_j$. For a validated child, let $F_{j,h}^{(k,\mathrm{child})}$ denote its average training fitness under budget $B$. Using the selected island's incumbent $F_{j,h-1}^{(k,\star)}$ and the current global incumbent $F_{j,h-1}^{(\mathrm{global},\star)}$ immediately before evaluation, define
\begin{equation}
\label{eq:island-reward}
\begin{aligned}
r_{j,h}^{(k)}
&=\operatorname{clip}\!\left(
\frac{F_{j,h}^{(k,\mathrm{child})}-F_{j,h-1}^{(k,\star)}}
{\max\!\left(F_{j,h-1}^{(\mathrm{global},\star)}
-F_{j,h-1}^{(k,\star)},0\right)+\epsilon},
0,1\right),\\
\delta_{j,h}^{(k)}
&=\max\!\left(
\frac{F_{j,h}^{(k,\mathrm{child})}-F_{j,h-1}^{(k,\star)}}
{\left|F_{j,h-1}^{(k,\star)}\right|+\epsilon},0\right).
\end{aligned}
\end{equation}
Here, $\operatorname{clip}(z,0,1)$ truncates $z$ to $[0,1]$, and $\epsilon>0$ prevents division by zero. The reward compares the child's gain over the island's incumbent with that island's gap to the current global incumbent, both measured before the update. For example, if these incumbents have fitness $80$ and $100$, respectively, a child with fitness $90$ receives reward $10/(20+\epsilon)\approx0.5$. A child that does not improve on the island's incumbent receives zero reward. If the island already matches the global incumbent, the denominator is $\epsilon$, so any gain of at least $\epsilon$ receives reward $1$. In contrast, $\delta$ measures nonnegative relative improvement within the island. These comparisons precede incumbent replacement. The Interpreter instead compares the child with its sampled parent.

\paragraph{Discounted scheduling updates.}
Each scheduled code-improvement visit updates the selected island's statistics:
\begin{equation}
\label{eq:island-updates}
\begin{aligned}
R_{j,h}^{(k)}
&=\rho R_{j,h-1}^{(k)}+r_{j,h}^{(k)},
&\qquad V_{j,h}^{(k)}
&=\rho V_{j,h-1}^{(k)}+1,\\
H_{j,h}^{(k)}
&=\rho H_{j,h-1}^{(k)}+(1-\rho)\delta_{j,h}^{(k)},
& n_{j,h}^{(k)}
&=n_{j,h-1}^{(k)}+1.
\end{aligned}
\end{equation}
The factor $\rho\in[0,1)$ controls discounting, and $H$ is the recent-progress signal used for plan review. If no valid child is obtained, the same update is applied once with $r=\delta=0$, without fabricating a fitness or replacing the incumbent. Non-selected islands retain their statistics.

\paragraph{Plateau criterion and complete procedure.}
For the selected island, $H_{j-1}^{(k)}\geq\tau$ results in one ordinary code-improvement visit, whereas $H_{j-1}^{(k)}<\tau$ triggers a plan review. A proposal starts a new plan generation only after its seed is validated, followed by $W$ consecutive local visits without another island selection or threshold test. If seed generation fails, the old plan and programs remain active and the round proceeds with one ordinary code-improvement visit.

Algorithm~\ref{alg:evolution} makes this update order explicit. As in Algorithm~\ref{alg:initialization}, \emph{CodePlan} and \emph{ImproveCode} include evaluation and debugging internally and return a validated program and its measured fitness, or $(\bot,\bot)$ after bounded failure during evolution. The temporary proposal $\widetilde P$ does not replace the current plan until \emph{RestartIsland} succeeds.

\begingroup
\makeatletter
\def\theHALG@line{supp.\Alph{section}.\arabic{algorithm}.\arabic{ALG@line}}
\makeatother
\begin{algorithm}[t]
\caption{The Evolution Procedure of HeurEvo.}
\label{alg:evolution}
\small
\begin{algorithmic}[1]
\Require $K$ initialized islands, training set $\mathcal D$,
per-instance budget $B$, rounds $J$, local steps $W\geq1$,
threshold $\tau$, library $\mathcal S_0$, codebase archive $\mathcal A_0$,
and debugging limit $M_{\mathrm{dbg}}$.
\For{$j=1,\ldots,J$}
    \State Carry forward all islands' plans, populations, feedback, and statistics
    \State $k\leftarrow\mathsf{SelectIsland}(j),\quad L\leftarrow1$
    \Comment{Eq.~\ref{eq:ucb}}
    \State Initialize local state $h=0$ from the preceding round
    \If{$H_{j-1}^{(k)}<\tau$}
        \State $\widetilde P\leftarrow\mathsf{ImprovePlan}(k)$
        \State $(G_j^{(k,\mathrm{seed})},F_j^{(k,\mathrm{seed})})
        \leftarrow\mathsf{CodePlan}(\widetilde P,\mathcal S_{j-1},B)$
        \If{$G_j^{(k,\mathrm{seed})}\neq\bot$}
            \State $\mathsf{RestartIsland}
            (k,\widetilde P,G_j^{(k,\mathrm{seed})},F_j^{(k,\mathrm{seed})})$
            \State $L\leftarrow W$
        \EndIf
    \EndIf
    \For{$h=1,\ldots,L$}
        \State Refresh the global incumbent from the current island incumbents
        \State $(G_{j,h}^{(k,\mathrm{par})},\mathcal I_{j,h}^{(k)})
        \leftarrow\mathsf{SelectCode}
        (\mathcal C_{j,h-1}^{(k)},\mathcal A_{j,h-1})$
        \State $(G_{j,h}^{(k,\mathrm{child})},F_{j,h}^{(k,\mathrm{child})})
        \leftarrow\mathsf{ImproveCode}
        (k,G_{j,h}^{(k,\mathrm{par})},\mathcal I_{j,h}^{(k)})$
        \If{$G_{j,h}^{(k,\mathrm{child})}\neq\bot$}
            \State $\Phi_{j,h}^{(k)}
            \leftarrow\mathsf{Interpret}
            (G_{j,h}^{(k,\mathrm{par})},G_{j,h}^{(k,\mathrm{child})})$
            \State $\mathsf{UpdateState}
            (k,G_{j,h}^{(k,\mathrm{child})},
            F_{j,h}^{(k,\mathrm{child})},\Phi_{j,h}^{(k)})$
            \Comment{Eqs.~\ref{eq:island-reward}--\ref{eq:island-updates}}
        \Else
            \State Apply Eq.~\ref{eq:island-updates} with
            $r_{j,h}^{(k)}=\delta_{j,h}^{(k)}=0$
            \State Retain programs, incumbent, and feedback; add no program
        \EndIf
    \EndFor
    \State Record the selected island's final local state and archive at round $j$
    \State $N_j\leftarrow\sum_{\ell=1}^{K}n_j^{(\ell)}$
    \State $\mathcal R_j
    \leftarrow\mathsf{SelectComponents}(\mathcal S_{j-1})$
    \Comment{Eq.~\ref{eq:component-review}}
    \State $\mathcal S_j
    \leftarrow\mathsf{ImproveComponents}(\mathcal S_{j-1},\mathcal R_j)$
\EndFor
\State \Return $\displaystyle
\operatorname*{arg\,max}_{G\in\mathcal A_J}F(G)$
\end{algorithmic}
\end{algorithm}
\endgroup

\paragraph{Compact controller notation in the main text.}
The pseudocode in Section~\ref{app:evolution} suppresses round indices. Immediately before global round $j$, the compact variables correspond to
\[
k^\star=k_j,\quad
\bar r_k=\frac{R_{j-1}^{(k)}}{V_{j-1}^{(k)}},\quad
n_k=n_{j-1}^{(k)},\quad
n_{\mathrm{tot}}=N_{j-1},\quad
H_k=H_{j-1}^{(k)}.
\]
They express the same UCB selection and plateau criterion specified above, without changing the reward or update rules.

\paragraph{System functions and agent-response calls.}
System functions execute prescribed sampling, database, evaluation, and numerical-update rules. Agent-response calls construct prompts, obtain structured responses, and parse the requested outputs.

\begin{table}[!htbp]
\centering
\setlength{\belowcaptionskip}{5pt}
\caption{System functions and agent-response calls.}
\label{tab:operator-types}
\begin{tabularx}{\linewidth}{@{}>{\raggedright\arraybackslash}p{0.30\linewidth}>{\raggedright\arraybackslash}X@{}}
\toprule
\textbf{Call category} & \textbf{Operators} \\
\midrule
System functions & \emph{SelectIsland}, \emph{SelectCode}, \emph{RestartIsland}, \emph{UpdateState}, \emph{SelectComponents} \\
Agent-response calls & \emph{ImprovePlan}, \emph{CodePlan}, \emph{ImproveCode}, \emph{Interpret}, \emph{ImproveComponents} \\
\bottomrule
\end{tabularx}
\end{table}

\emph{CodePlan} and \emph{ImproveCode} additionally include system-controlled evaluation and debugging. Their agents propose source code, but the returned fitness is measured by execution rather than supplied by an agent. The internal evaluator also determines whether a candidate must be sent to the Debugger. No separate \emph{EvaluateAndDebug} call is required in the pseudocode.

\paragraph{Island selection (\emph{SelectIsland}).}
\emph{SelectIsland} computes the UCB scores using Eq.~\ref{eq:ucb} in the main text and returns the selected island index. The decision uses the statistics available before the current round.

For an ordinary code-improvement visit, local index $h=0$ refers to the programs, incumbents, and statistics carried from the preceding global round. After an accepted plan proposal, $h=0$ instead refers to the new seed and the quantities initialized by \emph{RestartIsland}.

The cumulative counts $n$ and $N$ retain their history across plan generations. They are distinct from $V$, the discounted exposure statistic that is reset when a new generation starts.

\paragraph{Plan proposal (\emph{ImprovePlan}).}
\emph{ImprovePlan} gathers the selected island's plan and programs, cross-island plan details, active component descriptions, and accumulated plan-level feedback. It passes them to the Plan Improver and parses the returned proposal.

The existing plan remains available while the proposal is being implemented and validated. A proposal does not replace the current plan before a valid seed is obtained. Plan equality is not an acceptance criterion.

\paragraph{Seed generation and validation (\emph{CodePlan}).}
\emph{CodePlan} passes the proposed plan, component descriptions, and execution
budget to the Coder. It evaluates the generated program and invokes the
Debugger when compilation, validity, or runtime checks fail.

A successful call returns
\[
\left(G^{(\mathrm{seed})},F^{(\mathrm{seed})}\right),
\]
where the program is the final validated implementation and the fitness comes
from its evaluation.

Initialization continues this process until a valid seed is obtained. During
plan evolution, the configured debugging limit applies. If debugging is
exhausted, \emph{CodePlan} returns $(\bot,\bot)$. The proposal is then
discarded, the island retains its previous plan and programs, and the round
resumes with one ordinary code-improvement visit.

A failed seed-generation call does not increment the code-improvement visit
count and does not itself trigger a scheduling-reward update.

\paragraph{Island restart (\emph{RestartIsland}).}
\emph{RestartIsland} is invoked only after a proposed plan produces a valid
seed. It installs the proposed plan and initializes its parent-eligible
collection and incumbent with that seed. This handling applies even when the
proposal has identical content to the previous plan or its seed has lower
fitness than the previous incumbent.

Programs from the previous generation remain in the codebase according to the
database's retention policy, but they are no longer eligible parents for the
new generation. Previous plan-level feedback remains archived, and a new
current-generation feedback pool is started.

The function resets $R,V,H$ to $0,1,0$, respectively, without resetting $n$ or
$N$. The seed establishes a new within-generation baseline and does not receive
an improvement reward against the retired incumbent.

The island then receives $W$ consecutive local code-improvement visits. Neither
island selection nor the plan-evolution threshold is reconsidered between these
visits.

\paragraph{Parent and inspiration selection (\emph{SelectCode}).}
\label{app:code-selection}
\emph{SelectCode} samples one parent from the selected island's current
parent-eligible collection and inspiration programs from the codebase. The
parent must belong to the current plan generation, while inspirations may
provide examples from other plans.

The input $\mathcal A$ identifies the complete codebase. The collection
$\mathcal A^{(k,\mathrm{ins})}$ identifies the programs available as inspirations
for the selected island under the configured database policy. The returned
$\mathcal I_{j,h}^{(k)}$ contains the programs actually sampled for that visit.

The reference backends use related but distinct mechanisms. OpenEvolve's
documented parent-selection mechanisms combine random island sampling, elite
selection, and fitness-weighted sampling. Its configuration supports both
high-performing and diverse programs as prompt examples
\citep{openevolve}. AdaEvolve's published sampling subroutine switches
between uniform parent selection with diverse inspirations and parent selection
from the highest-fitness quartile with high-fitness inspirations
\citep{cemri2026adaevolve}.

HeurEvo applies its current-plan eligibility restriction through this database
interface. Parent and inspiration selection are repeated before every local
update, so later visits can use validated programs produced earlier in the
same round.

\paragraph{Code generation and validation (\emph{ImproveCode}).}
\emph{ImproveCode} constructs the Code Improver's prompt from the sampled
programs, current plan, component library, and available code-level feedback. It
then evaluates the generated child and performs bounded debugging when
necessary.

A successful call returns the final validated child and its measured average
training fitness. If debugging produces a valid repaired or regenerated
program, that final program---not an earlier rejected version---is returned.

When no valid child is obtained, the call returns $(\bot,\bot)$. Intermediate
rejected versions are not inserted into the Program Database. All internal
generation and debugging attempts remain part of the same code-improvement
visit.

A valid child with degraded fitness is different from a rejected child. It can
still be interpreted and submitted to the database's ordinary admission and
retention procedure.

\paragraph{Program comparison (\emph{Interpret}).}
\emph{Interpret} retrieves the selected parent's evaluation record and the
final validated child's record. It supplies the source code and per-step
implementation feedback to the Interpreter and parses the four-part report.

When debugging has modified the child, the comparison uses the validated result
and the originally selected parent. No additional program evaluation is
required merely to construct this comparison.

An unrepaired child does not receive a normal parent--child performance
interpretation because no valid child fitness is available. Its diagnostics
have already been used by the Debugger.

\paragraph{Database insertion and updates (\emph{UpdateState}).}
\emph{UpdateState} computes the normalized reward and relative improvement for
a valid child using the incumbent fitnesses immediately before the child's
evaluation. It then applies Eqs.~\ref{eq:island-reward}--\ref{eq:island-updates}
in the main text. Incumbent replacement occurs after these comparisons.

The Interpreter and scheduler use different references. The Interpreter
compares the child with its sampled parent. Scheduling measures progress
against the selected island's incumbent, which need not be the sampled parent.

The validated child is submitted to the Program Database with its measured
fitness, island and plan-generation association, implementation measurements,
and Interpreter report. The database manages retention and parent eligibility.
Feedback is made available to subsequent calls, and measured component
contributions are appended to the appropriate arrays.

If \emph{ImproveCode} returns failure, the visit receives
\[
r_{j,h}^{(k)}=0,\qquad\delta_{j,h}^{(k)}=0.
\]
The main-text update rule is applied with these values, including one increment
of $n$. No numerical child fitness is fabricated, no program is inserted, and
the incumbent remains unchanged. The discounting rule reduces previously
accumulated positive reward when no new reward is obtained.

A failed visit inside a $W$-visit block does not terminate the block. The
procedure continues with the next local visit using the available
parent-eligible programs. Rejected attempts do not automatically contribute
zero-valued component observations, because scheduling rewards and measured step
contributions serve different purposes.

\paragraph{Component selection (\emph{SelectComponents}).}
\emph{SelectComponents} considers existing components that are both present
in the active library and used by the selected island's current plan. A
component also needs at least one contribution observation collected under its
current description.

The function computes the review probabilities defined in
Appendix~\ref{app:component-contribution} and returns $\mathcal R_j$. Existing
components absent from the selected plan remain unchanged during that round.
New-component suggestions are considered separately from the contribution-based
selection of existing components.

\paragraph{Component revision and deferred removal (\emph{ImproveComponents}).}
\emph{ImproveComponents} passes the selected components, contribution
histories, aggregated feedback, and addition suggestions to the Component
Improver. Selection requests a review rather than requiring an edit.

When a description is revised, only that component's contribution array is
cleared. Arrays for unchanged components are retained. New components begin
with empty arrays.

When a component is removed, its description is deleted from the active library.
Existing plans and stored programs are not immediately rewritten, and their
measured fitness values remain associated with the original programs.
Subsequent code-improvement prompts suggest bypassing the affected step with
zero allocated time. Any child implementing that suggestion must pass ordinary
validation.
Deletion or replacement of the recorded plan entry occurs when the Plan Improver next revises the affected plan. It must delete that entry or replace its component with an available alternative.

\paragraph{Completion of a global round.}
Local visits are processed sequentially. Each visit uses the latest
parent-eligible collection, codebase, incumbent, and feedback. The global
reference is refreshed from the current island incumbents before each child
evaluation.

After the final local visit, the selected island's resulting programs,
feedback, and statistics are recorded for round $j$. Non-selected islands
retain their corresponding quantities. Component review then produces the
library used by subsequent rounds.

The current global incumbent is the best program among the current island
plans. The archive-best program is the best retained program in the codebase,
which may belong to a retired generation. Algorithm~\ref{alg:evolution} returns
the latter.

\subsection{Component Contribution and Review}
\label{app:component-contribution}
\label{app:component-review}

\paragraph{Intermediate fitness.}
For a program with $m$ steps, let $y_i$ denote its average best-so-far training fitness after step $i$, with final fitness $y=y_m$. The recorded sequence satisfies
\[
y_1\leq y_2\leq\cdots\leq y_m=y.
\]
These are higher-is-better fitness values, rather than raw minimization objectives or gaps. Equality is allowed when a step does not improve the incumbent.

Let $y_g$ denote the current global incumbent immediately before evaluation. The first-step reference is the mean of the recorded first-step observations available to the contribution calculation,
\begin{equation}
\label{eq:first-step-reference}
\bar y_1=
\frac{1}{M_{\mathrm{first}}}
\sum_{q=1}^{M_{\mathrm{first}}}y_{q,1}.
\end{equation}
Here, $M_{\mathrm{first}}$ counts first-step observations and is distinct from the code-improvement visit count $N$.

\paragraph{Contributions in multi-step programs.}
For $m\geq2$, the observed contribution of step $i$ is
\begin{equation}
\label{eq:multi-step-contribution}
\Delta_i=
\begin{cases}
0,
& i=1,\;y-\bar y_1\leq0,\\[5pt]
\operatorname{clip}\!\left(
\dfrac{y_1-\bar y_1}{y-\bar y_1+\epsilon},0,1
\right),
& i=1,\;y-\bar y_1>0,\\[10pt]
\operatorname{clip}\!\left(
\dfrac{y_i-y_{i-1}}{y-y_{i-1}+\epsilon},0,1
\right),
& 2\leq i<m,\\[10pt]
\operatorname{clip}\!\left(
\dfrac{y_i-y_{i-1}}
{\max(y_g,y)-y_{i-1}+\epsilon},0,1
\right),
& i=m.
\end{cases}
\end{equation}
The first step is compared with the historical first-step reference. Intermediate steps are normalized by the remaining improvement obtained by the complete program. The final step is normalized by the remaining distance to the better of the program's final fitness and the pre-evaluation global incumbent. The score prioritizes component review rather than estimating a causal effect.

With full indices restored, the contribution of step $i$ in local update $h$ is denoted by $\Delta_{j,h,i}^{(k)}$.

\paragraph{Contributions in single-step programs.}
For $m=1$, the only step is also the final step. Setting $y_0=\bar y_1$, its contribution is
\begin{equation}
\label{eq:single-step-contribution}
\Delta_1=
\begin{cases}
0,
& \max(y_g,y_1)\leq\bar y_1,\\[7pt]
\operatorname{clip}\!\left(
\dfrac{y_1-\bar y_1}
{\max(y_g,y_1)-\bar y_1+\epsilon},0,1
\right),
& \max(y_g,y_1)>\bar y_1.
\end{cases}
\end{equation}
The special case assigns zero contribution when the historical reference is at least as large as both the program's outcome and the current global incumbent. Otherwise, the denominator is strictly positive.

\paragraph{Per-component contribution histories.}
Each component maintains an independent contribution array,
\[
\boldsymbol{\Delta}_j(s)
=\left[\Delta_s^{[1]},\ldots,\Delta_s^{[M_j(s)]}\right].
\]
The array contains observations collected since that component's most recent description change. Each measured occurrence contributes one observation. A component used in several steps can therefore receive several numerical observations from one program, while its qualitative Interpreter feedback is consolidated into one component-level record.

For a nonempty array,
\begin{equation}
\label{eq:component-quality}
Q_j(s)=\frac{1}{M_j(s)}\sum_{q=1}^{M_j(s)}\Delta_s^{[q]}.
\end{equation}
A description revision clears the corresponding array. Observations associated with the previous description do not enter the revised component's numerical average. Historical Interpreter reports can remain available as contextual evidence. New components begin with empty arrays, and unchanged components retain their existing observations.

\paragraph{Review eligibility and probability.}
An existing component is eligible for review when it remains in the active library, appears in the selected island's current plan, and has at least one contribution observation under its current description. Thus,
\[
\mathcal R_j\subseteq
\left\{s\in\mathcal S_{j-1}:\;
s\text{ is used by }P_j^{(k_j)}, \,
M_j(s)\geq1
\right\}.
\]
For an eligible component, the review probability is
\begin{equation}
\label{eq:component-review}
p_j(s)=I_{\min}+
\frac{I_{\max}-I_{\min}}{1+C_I\sqrt{Q_j(s)}},
\end{equation}
where $0\leq I_{\min}\leq I_{\max}\leq1$ and $C_I>0$. A lower contribution score increases the probability of review without making revision mandatory.

There is no requirement to accumulate $W$ observations before reviewing a component. The $W$ consecutive code-improvement visits after an accepted plan proposal provide opportunities to gather multiple observations under unchanged descriptions, but they do not impose a minimum history length. After a description revision, one new measured contribution is sufficient for eligibility.

Unrepaired visits do not produce successful step measurements and therefore do not guarantee additional contribution observations. Suggestions for adding new components are assessed separately from the numerical review of existing components.

\section{Prompt Templates}
\label{app:prompts}

This appendix presents the prompt templates for initial solver generation and
plan-aware code evolution, together with their shared task context and
solution-reporting requirements. Fields in braces are filled with task-specific
information, optimization plans, and evaluation feedback.

\subsection{Task context and substitution fields}
\label{app:prompt-context}

The shared \texttt{problem\_context} contains the natural-language problem
description, a runtime command and parameter values, CSV table names and
locations, column descriptions, the mathematical model, and the reference
Gurobi solver code. The reference
implementation is authoritative when it disagrees with a textual description.
The first selected training instance supplies the representative prompt context.
The resulting program is evaluated on the selected training instances.
The table descriptions identify the data that the program must read
at execution time; CSV rows are not embedded in the prompt. Runtime parameter values
must be parsed by the generated program and can vary across executions.

Fields in braces are replaced when a message is constructed. Their meanings are:
\begin{itemize}[topsep=3pt,itemsep=0pt,parsep=0pt,partopsep=0pt,leftmargin=15pt]
  \raggedright
  \item \texttt{problem\_context}: the shared task context described above.
  \item \texttt{time\_limit}: the configured per-instance runtime allowance in
  seconds.
  \item \texttt{formatted\_plan} and \texttt{strategy\_context}: the validated
  runtime-aware plan and the descriptions of the strategy families
  selected by that plan. In the evolution prompt, \texttt{plan} provides the
  island's fixed strategy and subgoal for each step, and
  \texttt{selected\_strategies} supplies the corresponding strategy descriptions.
  \item \texttt{solution\_output\_requirements}: the task-specific contract for a
  complete solution report that can be checked by the verifier.
  \texttt{intermediate\_output\_requirements} inserts the shared reporting contract
  reproduced in Section~\ref{app:prompt-output}.
  \item \texttt{metrics}, \texttt{fitness\_description},
  \texttt{improvement\_areas}, and \texttt{artifacts}: the parent program's
  recorded performance, the scoring rule, comparisons with previous attempts,
  and execution feedback.
  \item \texttt{previous\_attempts}, \texttt{other\_context\_programs}, and
  \texttt{search\_guidance}: the selected ancestor history, top-performing and
  inspiration programs, and available guidance from prior evaluations.
  \item \texttt{current\_program} and \texttt{language}: the parent source code
  and its language. \texttt{timeout\_warning} reiterates the configured runtime
  allowance and the requirement to preserve the program's output format.
\end{itemize}

Instance-specific models, program histories, plans, and output requirements are
represented by placeholders so that the templates remain reusable across tasks.

\subsection{Shared intermediate-report contract}
\label{app:prompt-output}

Both user prompts require a complete solution report after each plan step and
at termination. The following instructions specify the shared reporting format
and cumulative elapsed-time measurements.

\promptfile{Shared intermediate-report contract}{prompts/intermediate_output_requirements.txt}

\subsection{Initial program generation}
\label{app:prompt-initial}

Unlike subsequent code evolution, initial program generation follows the supplied annotations and runtime fields and is instructed to implement every plan step. The system message defines the agent's role, and the user message supplies the task, plan, and implementation requirements.

\promptfile{Initial-generation system message}{prompts/initial_generation_system.txt}

\promptfile{Initial-generation user-message template}{prompts/runtime_coder_user_message.txt}

\subsection{Plan-aware code evolution}
\label{app:prompt-evolution}

The code-evolution prompt requests targeted SEARCH/REPLACE edits under a fixed recorded plan. It preserves step identifiers, component assignments, and subgoals while allowing implementation changes, runtime reallocation, and step deactivation through zero-time allocation. The experimental template is reproduced below with its original field names. The edits must preserve the plan's strategies and subgoals, respect the runtime allowance, and retain the required solution-reporting format.

\promptfile{Code-evolution system message}{prompts/code_evolution_system.txt}

\promptfile{Plan-aware code-evolution user-message template}{prompts/plan_based_diff_user_message.txt}

\end{document}